\documentclass[10pt,twocolumn,letterpaper]{article}

\usepackage{cvpr}              % To produce the CAMERA-READY version
\usepackage[dvipsnames, table]{xcolor}

\usepackage{graphicx}
\usepackage{amsmath}
\usepackage{amssymb}
\usepackage{booktabs}
\usepackage{mathtools}
\usepackage[accsupp]{axessibility}  % Improves PDF readability for those with disabilities.
\usepackage{color}
\usepackage{array}
\usepackage{pdflscape}
\usepackage[longtable]{multirow}
\usepackage{longtable}
\usepackage{tabularray}
\usepackage{booktabs}
\usepackage{times}
\usepackage{epsfig}
\usepackage{amsmath}
\usepackage{bbm}
\DeclareMathAlphabet\mathbfcal{OMS}{cmsy}{b}{n}
\usepackage{makecell}
\usepackage{siunitx}
\usepackage{svg}
\usepackage{float}
\usepackage{comment}
\usepackage{lipsum}
\usepackage{stfloats}
\usepackage{multicol}
\usepackage{multirow}
\usepackage{bm}
\usepackage{etoolbox}
\usepackage{icomma}
\usepackage{array}
\usepackage{tabulary}
\usepackage{xcolor}
\usepackage{paralist}
\usepackage{booktabs}
\usepackage{adjustbox}
\usepackage{caption}
\usepackage{subcaption}
\usepackage{arydshln}
\usepackage{rotating}

\definecolor{gray}{rgb}{0.3,0.3,0.3}
\definecolor{lightgray}{rgb}{0.8,0.8,0.8}
\definecolor{blue}{rgb}{0,0.5,1}
\definecolor{mask_red}{rgb}{1,0,0.8}
\definecolor{green}{rgb}{0.2,1,0.2}
\definecolor{rblue}{rgb}{0,0,1}
\definecolor{lightblue}{HTML}{6495ed}
\definecolor{lightred}{HTML}{F19C99}

\newcommand{\lightgray}[1]{\textcolor{lightgray}{#1}}

\definecolor{graytablerow}{gray}{0.6}

\usepackage{pifont}% http://ctan.org/pkg/pifont
\usepackage{tabu}
\usepackage[pro]{fontawesome5}

\usepackage{tikz}
\newcommand*\circled[1]{\tikz[baseline=(char.base)]{
\node[shape=circle,fill=gray,inner sep=0.598pt] (char) {\textcolor{white}{\small \textbf{#1}}};}}

\definecolor{cvprblue}{rgb}{0.21,0.49,0.74}
\usepackage[pagebackref,breaklinks,colorlinks,allcolors=cvprblue]{hyperref}
\usepackage[accsupp]{axessibility}  % Improves PDF readability for those with disabilities

\def\eg{\emph{e.g}\onedot} 

\def\ie{\emph{i.e}\onedot}

\def\wrt{w.r.t\onedot} 

\definecolor{MyDarkRed}{rgb}{0.8,0.02,0.02}
\definecolor{MyDarkBlue}{rgb}{0.02,0.02,0.8}
\definecolor{MyDarkGreen}{rgb}{0.1,0.8,0.1}
\definecolor{darkgreen}{rgb}{0.0, 0.5, 0.0}

\title{Scene-agnostic Pose Regression for Visual Localization 
}

\author{
Junwei Zheng$^1$ \quad 
Ruiping Liu$^1$ \quad 
Yufan Chen$^1$ \quad 
Zhenfang Chen$^4$ \quad 
Kailun Yang$^3$\\Jiaming Zhang$^{1,2,*}$ \quad 
Rainer Stiefelhagen$^1$\\
\normalsize
$^1$Karlsruhe Institute of Technology
\normalsize \quad
$^2$ETH Zurich
\normalsize \quad
$^3$Hunan University
\normalsize \quad
$^4$MIT-IBM Watson AI Lab
}

\begin{document}

%%%%%%%% Teaser figure
\twocolumn[{%
\renewcommand\twocolumn[1][]{#1}%
\maketitle
\begin{center}
    \centering
    \captionsetup{type=figure}
    \includegraphics[width=1.0\textwidth]{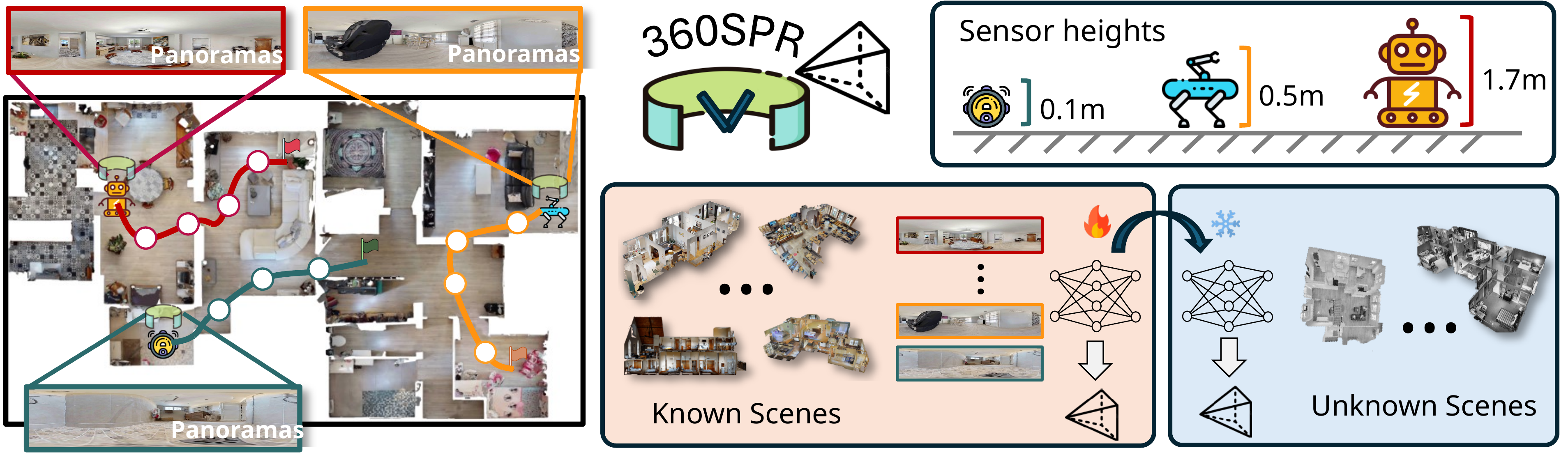}
    % \vskip -2ex
    \captionof{figure}{\textbf{360SPR} panoramic dataset for Scene-agnostic Pose Regression (SPR).
    Trajectories with different lengths and sampling intervals are collected at 3 heights of 360{\textdegree} sensors, \ie, $0.1{m}$, $0.5{m}$, $1.7{m}$, corresponding to sweeping~{\includegraphics[width=3mm]{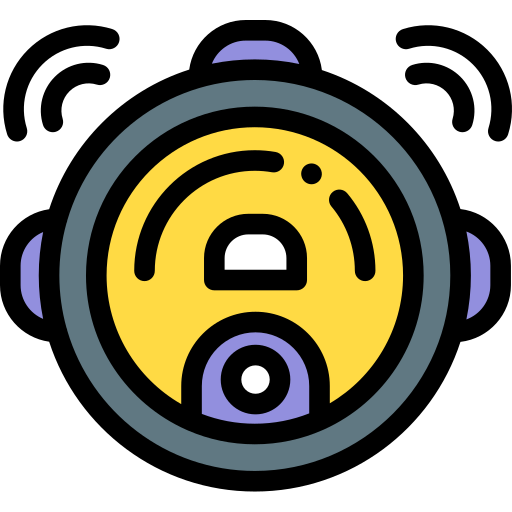}, quadruped~\includegraphics[width=3mm]{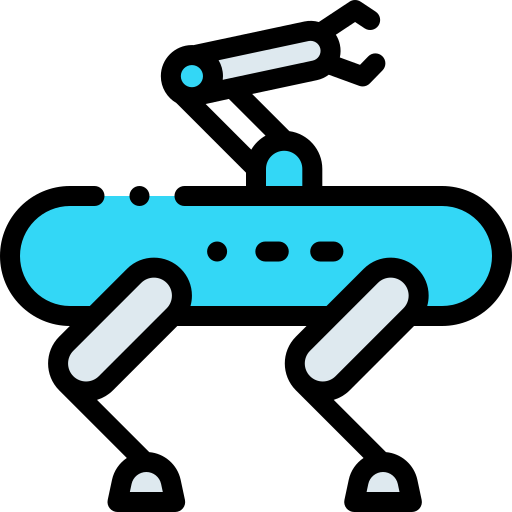}, and humanoid~\includegraphics[width=3mm]{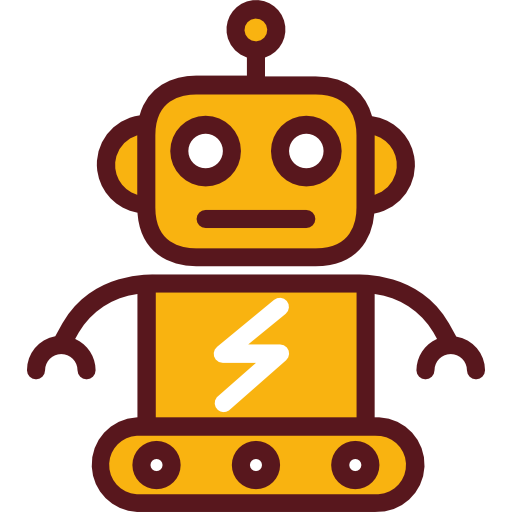}} robots. 
    The proposed SPR model trained from known scenes can generalize well to unknown scenes, without database retrieval.
    }
    \label{fig:360spr}
\end{center}%
}]

{
  \renewcommand{\thefootnote}
    {\fnsymbol{footnote}}
  \footnotetext[1]{Corresponding author (e-mail: {\tt jiaming.zhang@kit.edu}).}
}
\begin{abstract}
\label{sec:abstract}
Absolute Pose Regression (APR) predicts 6D camera poses but lacks the adaptability to unknown environments without retraining, while Relative Pose Regression (RPR) generalizes better yet requires a large image retrieval database. Visual Odometry (VO) generalizes well in unseen environments but suffers from accumulated error in open trajectories. To address this dilemma, we introduce a new task, \textbf{Scene-agnostic Pose Regression (SPR)}, which can achieve accurate pose regression in a flexible way while eliminating the need for retraining or databases. To benchmark SPR, we created a large-scale dataset, \textbf{360SPR}, with over $200{K}$ photorealistic panoramas, $3.6{M}$ pinhole images and camera poses in $270$ scenes at three different sensor heights. Furthermore, a \textbf{SPR-Mamba} model is initially proposed to address SPR in a dual-branch manner. Extensive experiments and studies demonstrate the effectiveness of our SPR paradigm, dataset, and model. In the unknown scenes of both 360SPR and 360Loc datasets, our method consistently outperforms APR, RPR and VO. The dataset and code are available at \href{https://junweizheng93.github.io/publications/SPR/SPR.html}{SPR}.
\end{abstract}

\section{Introduction}
\label{sec:intro}
Visual localization is one of the fundamental tasks in the computer vision domain.
It is required to predict the 6D camera poses given the visual cues captured by a camera.
The ability to determine the camera poses from captured images enables various downstream applications that require precise spatial awareness, \eg, VR/AR~\cite{illahi2023RRVR,sarlin2022lamar}, autonomous driving~\cite{moreau2022coordinet}, and robotics~\cite{qiao2023transapr,ruan2023combining}. 

Absolute Pose Regression (APR) is one of the classical paradigms for camera pose regression.
Given the images of a scene, the model predicts the absolute camera poses concerning the scene coordinate system.
Since a model in the APR paradigm only learns the scene-specific features, it is not applicable in unknown environments without retraining.
Relative Pose Regression (RPR), in contrast, generalizes better in unknown environments.
This paradigm learns the relative pose between a reference and query image during training.
In the inference phase, the model first retrieves a reference image from the training set and then predicts the relative pose for a query image. Although RPR is able to predict camera poses in unseen environments, it requires a large database to retrieve reference images similar to the query image, which plays an important role in the whole RPR paradigm.
Without enough overlap or similarity between reference and query images, the model performance of RPR drops dramatically.
Visual Odometry (VO) predicts the next camera pose based on the previously predicted pose and can generalize well in unseen environments.
However, it suffers from accumulated drift in open trajectories.

To address the dilemma of APR, RPR and VO, in this work, we introduce a novel task termed \textbf{Scene-agnostic Pose Regression (SPR)}, targeting the generalization problem and the need for a large database.
Given a sequence of images along a trajectory, SPR takes the first image as the coordinate system origin.
Using all images before the query image along the path as the model input, the camera pose of the query image is predicted \textit{\wrt} the origin image, not the previous one.
It's worth noting that there is no accumulated drift for open trajectories in the SPR paradigm compared to VO since the prediction of the current frame camera pose doesn't depend on the previous frame camera pose.
Compared with APR, SPR disentangles the coordinate system from scenes and learns the relative poses between frames instead of scene-specific features so it is applicable in unseen environments.
Unlike RPR, SPR bypasses the need to use a large database for image retrieval.
The relative poses are calculated between the first frame of the trajectory and the desired query frames.

\begin{figure}[tb]
    \centering
    \includegraphics[width=\linewidth]{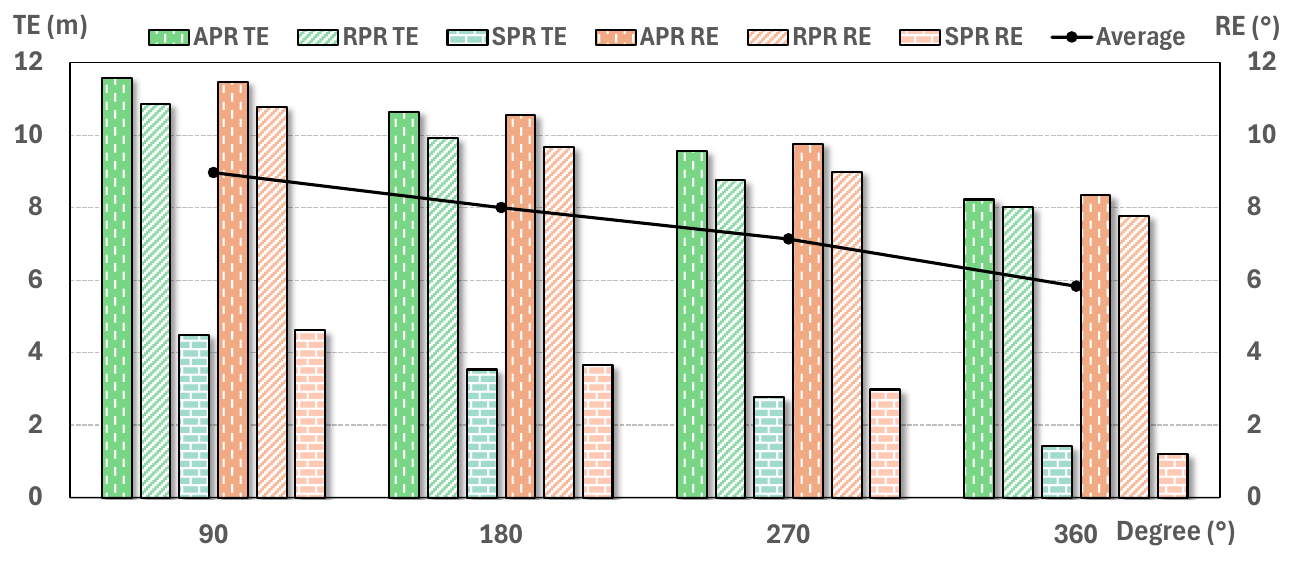}
    % \vskip -2ex
    \caption{Model performance in APR, RPR, and SPR paradigms with the change of image field of view. TE and RE stand for Translation Error and Rotation Error, respectively.}
    \label{fig:fov_comparison}
    \vskip -2ex
\end{figure}

Most publicly available visual localization datasets utilize pinhole images to perform camera pose regression while only a small amount of datasets use panoramas, \eg, the 360Loc~\cite{huang2024360loc} dataset.
Fig.~\ref{fig:fov_comparison} illustrates the average median translation and rotation error of APR (represented by PoseNet~\cite{kendall2015posenet}), RPR (represented by Relpose-GNN~\cite{turkoglu2021relpose}) and SPR (represented by SPR-Mamba) with the change of image Field of View (FoV) in known environments of the 360Loc dataset.
Both translation and rotation errors decrease when the image FoV increases, proving the necessity of utilizing panoramas in pose regression tasks.
The reason behind this phenomenon is straightforward.
Panoramas~\cite{zheng2024ops,hu2024deformable,wei2024onebev} not only enrich necessary visual information for camera pose regression but also enable sufficient overlap and similarity between frames, which is extremely important for RPR and SPR.
The limitation of the panoramic visual localization dataset, 360Loc~\cite{huang2024360loc}, is also obvious because it only provides less than $10{K}$ panoramas distributed in only $4$ scenes with a fixed sampling interval and sensor height.
The limited data capacity and diversity cannot meet the demand for accurate and robust spatial awareness in various real-world applications.
As shown in Table~\ref{tab:dataset_cross-evaluation}, the model trained on 360Loc~\cite{huang2024360loc} is not able to perform well in other diverse environments with an error increase of $4.08{m}/4.67{^\circ}{\uparrow}$ on translation and rotation, respectively.
Table~\ref{tab:height_cross-evaluation} further demonstrates the lack of robustness when the model is trained at a fixed sensor height.
To this end, we establish a large-scale dataset named \textbf{360SPR} with over $200{K}$ photorealistic panoramas and $3.6{M}$ pinhole images with camera poses in $270$ scenes using the Habitat simulator~\cite{savva2019habitat1,szot2021habitat2,puig2023habitat3} powered by HM3D~\cite{ramakrishnan2021hm3d} and Matterport3D~\cite{chang2017matterport3d} datasets.
Fig.~\ref{fig:360spr} illustrates the data collection process.
Trajectories of different lengths, with varying sampling intervals between sampling points along the path, are collected at three different robot heights, \ie, $1.7{m}$, $0.5{m}$, $0.1{m}$, corresponding to humanoid robots~\includegraphics[width=3mm]{figures/icon_robot.png}, quadruped robots~\includegraphics[width=3mm]{figures/icon_dog.png}, and sweeping robots~\includegraphics[width=3mm]{figures/icon_vacuum.png}, respectively.

Apart from the SPR task and 360SPR dataset, we propose a new model termed \textbf{SPR-Mamba} to explore the effectiveness of the SPR paradigm.
SPR-Mamba consists of a local branch and a global branch.
The local branch learns the relative pose between the current and previous frame while the global branch focuses on the one between the query and origin frame.
These complementary branches enable comprehensive learning for camera pose regression.
Extensive experiments on the 360SPR and 360Loc~\cite{huang2024360loc} benchmarks verify the effectiveness of the SPR with an error decrease over $\bf{7{m}/16{^\circ}}$ on translation and rotation in unseen environments, outperforming APR and RPR paradigms in the camera pose regression task.

To summarize, our contributions are as follows: 
\begin{compactitem}
    \item We propose a new task termed \textbf{Scene-agnostic Pose Regression (SPR)}, addressing the generalization problem of APR, the demand for a large database of RPR and the accumulated error of VO.
    \item We create \textbf{360SPR}, a dataset with $200{K}$ panoramas and $3.6{M}$ pinhole images across $270$ scenes at three robot heights for panoramic visual localization.
    \item We introduce a model termed \textbf{SPR-Mamba} consisting of a local and global branch. The local branch learns the relative pose between the current and the previous image while the global branch focuses on the one between the query and the origin image.
    \item Extensive experiments on 360SPR and 360Loc~\cite{huang2024360loc} datasets prove the effectiveness of the proposed SPR paradigm and SPR-Mamba model with an error reduction over $\bf{7{m}/16{^\circ}}$ on translation and rotation in unseen environments compared to APR and RPR.
\end{compactitem}

\begin{figure*}[tbh]
    \centering
    \includegraphics[width=0.99\linewidth]{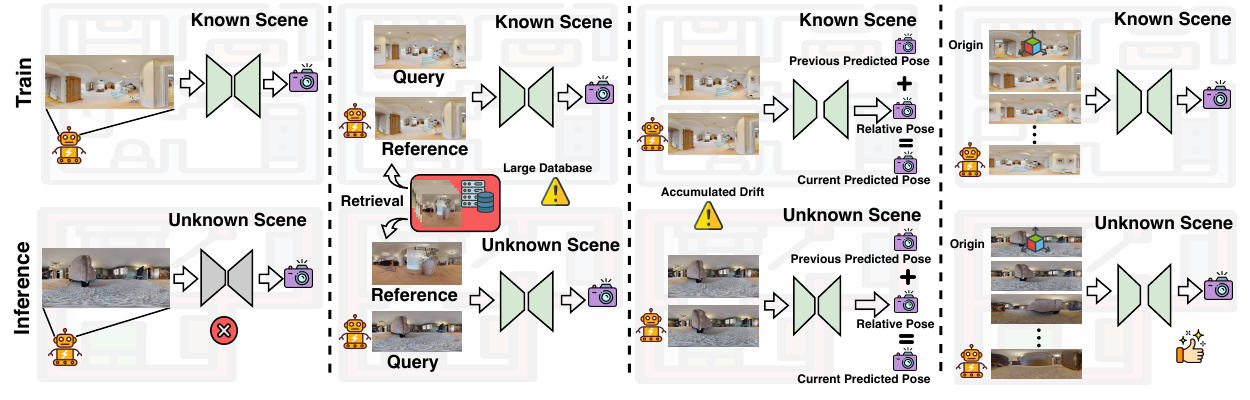}
    \begin{subfigure}[t]{0.25\textwidth}
        \centering
        \vskip-3ex
        \caption{Absolute Pose Regression}\label{fig3-a}
    \end{subfigure}%\hfill
    \begin{subfigure}[t]{0.25\textwidth}
        \centering
        \vskip-3ex
        \caption{Relative Pose Regression}\label{fig3-b}
    \end{subfigure}%\hfill
     \begin{subfigure}[t]{0.25\textwidth}
        \centering
        \vskip-3ex
        \caption{Visual Odometry} \label{fig3-c}
    \end{subfigure}%\hfill
     \begin{subfigure}[t]{0.25\textwidth}
        \centering
        \vskip-3ex
        \caption{Scene-agnostic Pose Regression} \label{fig3-d}
    \end{subfigure}%\hfill
        \vskip-2ex
    \caption{Comparison of different paradigms. 
    (a) APR is not applicable in unseen scenes without retraining. 
    (b) RPR requires a large database for image retrieval. 
    (c) VO suffers from accumulated drift.
    (d) SPR (ours) sets the first image in a sequence as the origin and predicts the relative pose of its followings, \ie, query images, avoiding the need of retraining, retrieval database or accumulated drift.
    }
    \vskip -2ex
    \label{fig:spr}
\end{figure*}

\section{Related Work}
\label{sec:related_work}
\noindent\textbf{Absolute Pose Regression (APR)} methods~\cite{kendall2015posenet,shavit2021ms-transformer,saha2018anchorpoint,chen2024marepo,lin2024pmnet,clark2017vidloc,brahmbhatt2018mapnet,xue2019lsg,xue2020gl-net,kendall2016modelling,wu2017delving,kendall2017geometric} aim to directly predict the 6D camera poses from images within a known environment used for training. 
PoseNet~\cite{kendall2015posenet} employed a CNN to regress poses from single images. MS-Transformer~\cite{shavit2021ms-transformer} utilized transformer~\cite{vaswani2017transformer} for multiple scenes. 
AnchorPoint~\cite{saha2018anchorpoint} introduced anchor points to enhance localization, while Marepo~\cite{chen2024marepo} conditioned the pose regressor on a scene-specific map representation and PMNet~\cite{lin2024pmnet} leveraged neural volumetric pose features for localization. 
Besides, sequence-based APR methods leverage temporal dependency information from image sequences to improve absolute pose prediction performance. 
VidLoc~\cite{clark2017vidloc} extended PoseNet by incorporating long short-term memory networks to capture temporal dependencies. MapNet~\cite{brahmbhatt2018mapnet}, LS-G~\cite{xue2019lsg}, and GL-Net~\cite{xue2020gl-net} integrated geometric consistency and global-local feature fusion. While APR requires retraining on new scenes due to their scene-specific features, 
our SPR predicts camera poses in unknown scenes without retraining or dependence on scene-specific maps. 

\noindent\textbf{Relative Pose Regression (RPR)} methods~\cite{laskar2017nn-net,balntas2018relocnet,ding2019camnet,yang2019extreme,li2019line,zhou2020essnet,turkoglu2021relpose,yang2020rcpnet,zhang2022relative,guan2021relative,sarlin202pixloc} estimate the 6D camera relative pose between the query image and reference image, generalizing to unseen scenes compared to APR methods. 
NN-Net~\cite{laskar2017nn-net} retrieved similar database images and predicted relative poses without requiring scene-specific training. 
CamNet~\cite{ding2019camnet} presented a coarse-to-fine retrieval framework that enhances retrieval accuracy. RelocNet~\cite{balntas2018relocnet} learned feature embeddings through continuous metric learning. EssNet~\cite{zhou2020essnet} directly predicted the essential matrix from image pairs, facilitating robust RPR. RelPose-GNN~\cite{turkoglu2021relpose} utilized a GNN to model spatial relationships between cameras. Although RPR methods demonstrate better generalization than APR methods in unseen environments, they normally require a large database for image retrieval and feature matching, which can be impractical for large-scale or dynamic settings. Additionally, the pose regression results may suffer from limited accuracy due to challenges in matching and environmental variations. Unlike RPR, SPR eliminates the reliance on large-scale reference image databases, yielding better pose prediction in unseen cases.

\noindent\textbf{Visual Odometry (VO)} is a core technique~\cite{dai2022self,sun2022improving,wagstaff2020self,li2020self,zou2020learning,li2021generalizing,messikommer2024reinforcement,yu2023robust,morra2023mixo,shen2023dytanvo,naumann2024nerf,teed2024deep} used in Simultaneous Localization and Mapping (SLAM)~\cite{cao2023tightly,morelli2023colmap} to estimate the trajectory of a camera by analyzing sequential image data to facilitate navigation in an unknown environment.
Traditional feature-based VO methods~\cite{krombach2016combining, krombach2018feature, engel2013semi, gladkova2021tight, badino2013visual} rely on identifying and tracking specific image features to estimate camera movement. 
Chien~\textit{et al.}~\cite{chien2016feature} evaluated the choice of different features for monocular visual odometry. 
Learning-based VO methods leverage deep learning for visual localization. DeepVO~\cite{wang2017deepvo} utilized neural networks to predict trajectories from image sequences, while D3VO~\cite{yang2020d3vo} further added depth data for better accuracy. VRVO~\cite{zhang2022vrvo} was tailored for the smooth tracking demands of virtual environments. DAVO~\cite{kuo2020davo} was designed to quickly adapt to dynamic settings. CEGVO~\cite{ji2024cegvo} employed a novel loss function to enhance visual odometry. ColVO~\cite{liu2024colvo} was proposed to estimate colon depth and colonoscopic pose continuously.
The open-loop trajectory of VO has unavoidable drift accumulated over time and distance. 
In contrast, SPR is more robust in open- and close-loop trajectories.

\begin{table*}[t]
\tiny
\setlength{\abovecaptionskip}{0pt}
\setlength{\belowcaptionskip}{0pt}
\caption{Comparison of visual localization datasets. Differences include: photographic \includegraphics[width=3mm]{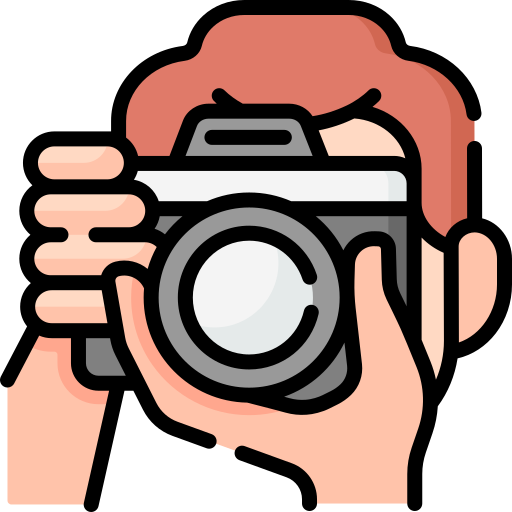} \textit{vs.} photorealistic \includegraphics[width=3mm]{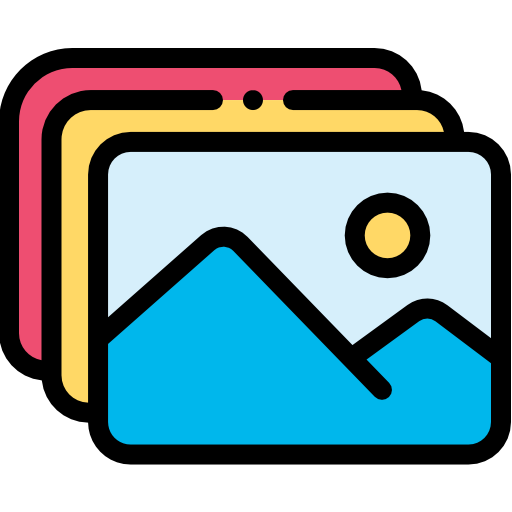} types; field of views (panoramic \textit{vs.} pinhole); image sizes; the number of scenes/trajectories/intervals; and sensor heights (in sweeping~\includegraphics[width=3mm]{figures/icon_vacuum.png}, quadruped~\includegraphics[width=3mm]{figures/icon_dog.png}, humanoid~\includegraphics[width=3mm]{figures/icon_robot.png} robots). 
Our 360SPR has $200{K}$ panoramas and $3.6{M}$ pinhole images, offering diverse data for visual localization.     
} 
\vskip -2ex
\label{tab:360spr}
\begin{center}
\resizebox{\linewidth}{!}{
\setlength{\tabcolsep}{1.6pt}
\begin{tabular}{lcccccccccc}
\toprule[1pt]
{\textbf{Dataset}}&{\textbf{Type}}&{\textbf{Panoramas}}&{\textbf{Pan. Res.}}&{\textbf{Pinholes}}&{\textbf{Pin. Res.}}&{\textbf{Scenes}}&{\textbf{Trajectories}}&{\textbf{Varying Interval}}&{\textbf{Height}}&{\textbf{Depth}}\\
\midrule
{7Scenes~\cite{glocker20137scenes}}&{\includegraphics[width=2mm]{figures/icon_photo.png}}&{\includegraphics[width=2mm]{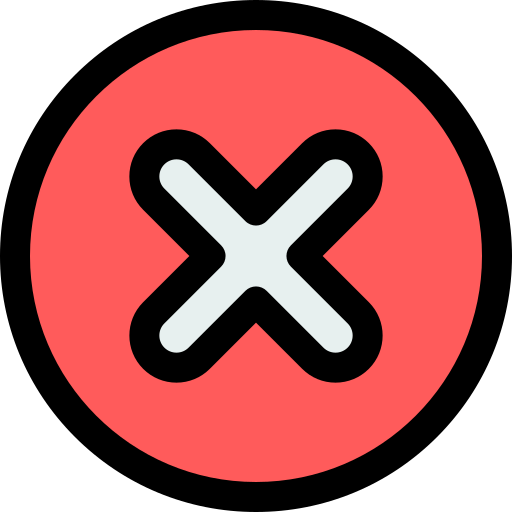}}&{\includegraphics[width=2mm]{figures/icon_crossmark.png}}&{33K}&{480$\times$640}&{7}&{\includegraphics[width=2mm]{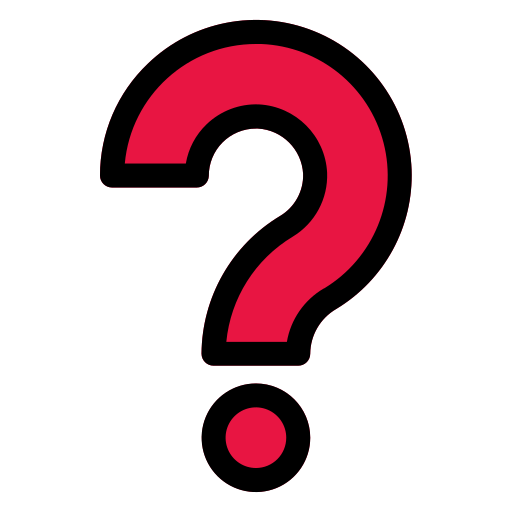}}&{\includegraphics[width=2mm]{figures/icon_questionmark.png}}&{\includegraphics[width=2mm]{figures/icon_robot.png}}&{\includegraphics[width=2mm]{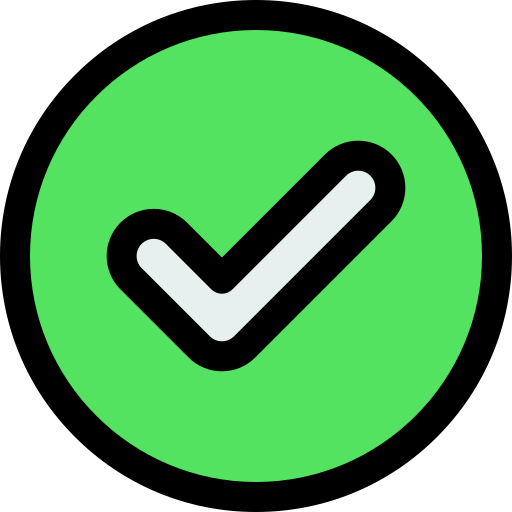}}\\
{12Scenes~\cite{valentin201612scenes}}&{\includegraphics[width=2mm]{figures/icon_photo.png}}&{\includegraphics[width=2mm]{figures/icon_crossmark.png}}&{\includegraphics[width=2mm]{figures/icon_crossmark.png}}&{246.7K}&{968$\times$1296}&{12}&{\includegraphics[width=2mm]{figures/icon_questionmark.png}}&{\includegraphics[width=2mm]{figures/icon_questionmark.png}}&{\includegraphics[width=2mm]{figures/icon_robot.png}}&{\includegraphics[width=2mm]{figures/icon_checkmark.png}}\\
{InLoc~\cite{taira2018inloc}}&{\includegraphics[width=2mm]{figures/icon_photo.png}}&{\includegraphics[width=2mm]{figures/icon_crossmark.png}}&{\includegraphics[width=2mm]{figures/icon_crossmark.png}}&{14K}&{1200$\times$1600}&{5}&{\includegraphics[width=2mm]{figures/icon_questionmark.png}}&{\includegraphics[width=2mm]{figures/icon_questionmark.png}}&{\includegraphics[width=2mm]{figures/icon_robot.png}}&{\includegraphics[width=2mm]{figures/icon_crossmark.png}}\\
{Cambridge~\cite{kendall2015posenet}}&{\includegraphics[width=2mm]{figures/icon_photo.png}}&{\includegraphics[width=2mm]{figures/icon_crossmark.png}}&{\includegraphics[width=2mm]{figures/icon_crossmark.png}}&{13.2K}&{1080$\times$1920}&{5}&{\includegraphics[width=2mm]{figures/icon_questionmark.png}}&{\includegraphics[width=2mm]{figures/icon_crossmark.png}}&{\includegraphics[width=2mm]{figures/icon_robot.png}}&{\includegraphics[width=2mm]{figures/icon_checkmark.png}}\\
{NaVIP~\cite{yu2024navip}}&{\includegraphics[width=2mm]{figures/icon_photo.png}}&{\includegraphics[width=2mm]{figures/icon_crossmark.png}}&{\includegraphics[width=2mm]{figures/icon_crossmark.png}}&{300K}&{\textbf{2160$\times$3840}}&{4}&{\includegraphics[width=2mm]{figures/icon_questionmark.png}}&{\includegraphics[width=2mm]{figures/icon_questionmark.png}}&{\includegraphics[width=2mm]{figures/icon_robot.png}}&{\includegraphics[width=2mm]{figures/icon_crossmark.png}}\\
{LaMAR~\cite{sarlin2022lamar}}&{\includegraphics[width=2mm]{figures/icon_photo.png}}&{\includegraphics[width=2mm]{figures/icon_crossmark.png}}&{\includegraphics[width=2mm]{figures/icon_crossmark.png}}&{152K}&{480$\times$640}&{3}&{\includegraphics[width=2mm]{figures/icon_questionmark.png}}&{\includegraphics[width=2mm]{figures/icon_questionmark.png}}&{\includegraphics[width=2mm]{figures/icon_robot.png}}&{\includegraphics[width=2mm]{figures/icon_checkmark.png}}\\
{360Loc~\cite{huang2024360loc}}&{\includegraphics[width=2mm]{figures/icon_photo.png}}&{9.3K}&{\textbf{3072$\times$6144}}&{14.2K}&{1200$\times$1920}&{4}&{18}&{\includegraphics[width=2mm]{figures/icon_crossmark.png}}&{\includegraphics[width=2mm]{figures/icon_robot.png}}&{\includegraphics[width=2mm]{figures/icon_checkmark.png}}\\
\midrule
{360SPR (ours)}&{\includegraphics[width=2mm]{figures/icon_picture.png}}&{\textbf{200K}}&{1024$\times$2048}&{\textbf{3.6M}}&{512$\times$512}&{\textbf{270}}&{\textbf{20K}}&{\includegraphics[width=2mm]{figures/icon_checkmark.png}}&{\includegraphics[width=2mm]{figures/icon_vacuum.png}~\includegraphics[width=2mm]{figures/icon_dog.png}~\includegraphics[width=2mm]{figures/icon_robot.png}}&{\includegraphics[width=2mm]{figures/icon_checkmark.png}}\\
\bottomrule
\end{tabular}
}
\end{center}
\end{table*}

\section{Methodology}
\label{sec:methodology}
The scene-agnostic Pose Regression (SPR) task is formulated in Sec.~\ref{subsec:spr} and compared with APR, RPR and VO.
Besides, our large-scale 360SPR dataset for benchmarking the SPR task is presented in Sec.~\ref{subsec:360spr}.
In Sec.~\ref{subsec:spr-mamba}, we introduce the proposed dual-branch SPR-Mamba model.

\subsection{Scene-agnostic Pose Regression}
\label{subsec:spr}
\noindent \textbf{Task Definition.}
The Scene-agnostic Pose Regression task aims to estimate the camera pose $\mathbf{T}_q$ of a query image $I_q$ relative to an origin image $I_1$ within an arbitrary scene, independent of specific scene characteristics or databases. Given a sequence of images ${I_1, I_2, \ldots, I_q}$ captured along a trajectory, the SPR task requires the model to accurately compute $\mathbf{T}_q$, which represents the camera's position and orientation for image $I_q$. Unlike traditional pose estimation tasks that rely on structured scenes or predefined databases, SPR operates across diverse environments, using only the image sequence itself without scene-specific priors. The goal of SPR is to develop a robust model that outputs $\mathbf{T}_q$ \textit{\wrt} $I_1$ by processing the entire sequence ${I_1, \ldots, I_{q-1}, I_q}$.

\noindent \textbf{Task Difference.} 
Fig.~\ref{fig:spr} showcases the difference between APR, RPR, VO and our proposed SPR task.
\circled{1} APR (in Fig.~\ref{fig3-a}) is capable of predicting camera poses in seen environments occurring in the training set.
However, it is not applicable in unseen environments during inference since the model learns scene-specific features in the APR paradigm.
\circled{2} RPR (in Fig.~\ref{fig3-b}), in contrast to APR, focuses on the relative features between image pairs during training.
Although RPR has better generalizability compared to APR, it requires a large database during inference since RPR needs to retrieve a reference image similar to the query image to form an image pair as the model input.
\circled{3} VO (in Fig.~\ref{fig3-c}) can generalize well in unseen environments but suffers from accumulated drift since it uses the previously predicted camera pose to predict the current pose.
\circled{4} SPR (in Fig.~\ref{fig3-d}) is proposed to address the aforementioned problems.
By choosing the first image in the sequence as the scene origin, SPR separates the coordinate system from particular scenes in the training set and identifies scene-agnostic features between the query and the origin image.
This enables the model to perform more effectively in novel scenes without the need for a large-scale database.
It is worth noting that open trajectories in the SPR paradigm do not experience accumulated drift, as the regression of the current camera pose is not influenced by the preceding pose.

\subsection{360SPR: Established Dataset}
\label{subsec:360spr}
\noindent \textbf{Data Collection.} 
We first collect pinhole images in size of $512{\times}512$ using the Habitat simulator~\cite{savva2019habitat1,szot2021habitat2,puig2023habitat3} powered by HM3D~\cite{ramakrishnan2021hm3d} and Matterport3D~\cite{chang2017matterport3d} datasets and then stitch the pinhole images to obtain panoramas in $1024{\times}2048$.
We use the same stitching tool as Matterport3D~\cite{chang2017matterport3d} dataset.
For every sample point in the trajectories, we collect images with $3$ elevations and $6$ headings, resulting in $18$ pinhole images with their corresponding camera poses.
We select the camera pose of the $10$-th pinhole image in the sequential sequence as the pose of the stitched panorama.
To enable high-quality panoramic images, three inspectors manually checked all samples in the form of cross validation. 
The whole cleaning process took more than $300$ hours.
Fig.~\ref{fig:360spr} showcases the data collection process.
Within a navigable area of a scene, we randomly select two points as the starting and destination points.
Then we calculate the shortest path between the two points using the Dijkstra~\cite{dijkstra2022dijkstra} algorithm.
Since the 360Loc~\cite{huang2024360loc} dataset doesn't consider different sampling intervals and sensor heights, it's difficult to satisfy the need for robust and accurate spatial awareness in various real-world applications.
To this end, we sample trajectories in different lengths with varying sampling intervals between sampling points along the path.
The trajectory length in 360SPR varies from $3{m}$ to $20{m}$ and the number of panoramas in one trajectory varies from $5$ to $20$.
Moreover, three different robot heights with a sampling ratio of $1{:}1{:}2$ are also taken into account, \ie, sweeping~{\includegraphics[width=3mm]{figures/icon_vacuum.png}, quadruped~\includegraphics[width=3mm]{figures/icon_dog.png}, and humanoid~\includegraphics[width=3mm]{figures/icon_robot.png}} robots as shown in Fig.~\ref{fig:360spr}.

\noindent \textbf{Data Statistics.} 
Table~\ref{tab:360spr} lists the statistics of different visual localization datasets, including 7Scenes~\cite{glocker20137scenes}, 12Scenes~\cite{valentin201612scenes}, InLoc~\cite{taira2018inloc}, Cambridge~\cite{kendall2015posenet}, NaVIP~\cite{yu2024navip}, LaMAR~\cite{sarlin2022lamar}, and 360Loc~\cite{huang2024360loc}. All these datasets are photographic. 
As 360SPR is collected using the Habitat~\cite{savva2019habitat1,szot2021habitat2,puig2023habitat3} simulator, the stitched panoramas are photorealistic rather than photographic, providing the potential for domain adaptation and generalization tasks. Besides, our 360SPR dataset focuses on indoor scenes with different sensor heights.  
Compared with other visual localization datasets, 360SPR showcases its value with more than $200{K}$ panoramas and $3.6{M}$ pinholes distributed in $20{K}$ trajectories among $270$ scenes, addressing the data capacity and diversity problem of existing datasets for visual localization.
For the visualization of the 360SPR panoramas, please refer to the supplementary.

\begin{figure*}[tbh]
    \centering
    \includegraphics[width=0.99\linewidth]{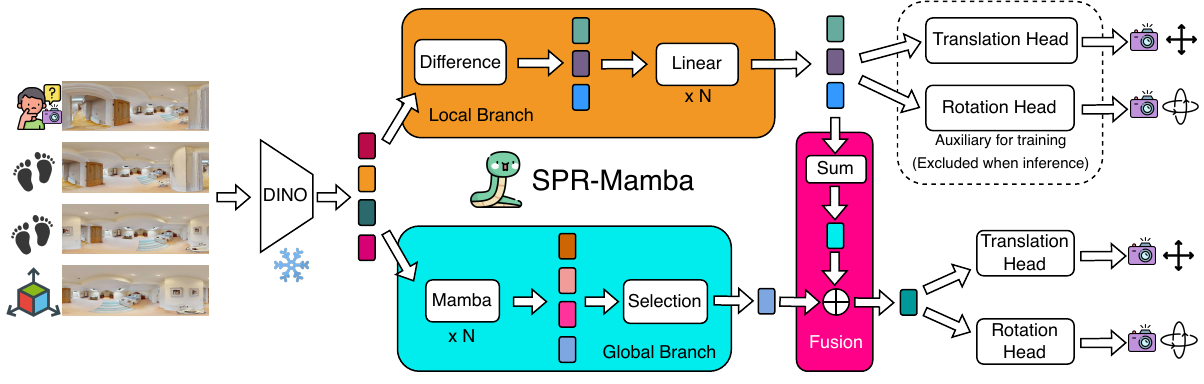}
    \vskip -2ex
    \caption{Architecture of SPR-Mamba. The local branch learns the relative poses between adjacent frames while the global branch focuses on the pose between the query and the first frame in the sequence. The auxiliary heads for the local branch are not necessary after training. Note that SPR-Mamba can handle arbitrary sequence length during inference and we use $4$ images for the demonstration purpose.}
    \label{fig:spr-mamba}
    \vskip -2ex
\end{figure*}

\subsection{SPR-Mamba Framework}
\label{subsec:spr-mamba}
\noindent \textbf{Preliminary.} 
State Space Models~\cite{gu2021ssm} (SSMs) and Mamba~\cite{mamba,mamba2} have been receiving increasing interest in the community due to their excellent performance in modeling sequential inputs and outputs.
SSMs can be expressed by using a hidden state $h(t) \in \mathbb{R}^N$, parameter $\mathbf{A}\in \mathbb{R}^{N\times N}$, $\mathbf{B} \in \mathbb{R}^{N}, \mathbf{C}\in \mathbb{R}^{N}$ as follows:
\begin{align}
    h^{\prime}(t) &= \mathbf{A}h(t) + \mathbf{B}x(t), \label{eq:continuous_ssm_hidden_state} \\
    y(t) &= \mathbf{C}h(t). \label{eq:continuous_ssm_output}
\end{align}
Discretizing SSMs~\cite{gu2021ssm} by leveraging a timescale parameter $\boldsymbol{\Delta}$, Equations~\ref{eq:continuous_ssm_hidden_state} and \ref{eq:continuous_ssm_output} can be reformulated as:
\begin{align}
    h_t &= \bar{\mathbf{A}}h_{t-1} + \bar{\mathbf{B}}x_t,\\ 
    y_t &= \mathbf{C}h_t,
\end{align}
where
\begin{align}
    \bar{\mathbf{A}} &= \exp\left( \boldsymbol{\Delta} \mathbf{A} \right),\\
    \bar{\mathbf{B}} &= \left( \boldsymbol{\Delta} \mathbf{A} \right)^{-1} \left( \exp \left( \boldsymbol{\Delta} \mathbf{A} - \mathbf{I} \right) \right) \cdot \boldsymbol{\Delta} \mathbf{B}.
\end{align}
Compared to RNNs and Transformers~\cite{vaswani2017transformer}, one advantage of SSMs is that they can be trained in parallel while doing inference in a sequential manner: 
\begin{align}
    \bar{\mathbf{K}} &= \left( {\mathbf{C}} \bar{\mathbf{B}}, {\mathbf{C}} \bar{\mathbf{A}} \bar{\mathbf{B}}, \dots, {\mathbf{C}} \bar{\mathbf{A}}^{L-1} \bar{\mathbf{B}} \right),\\
    y &= x \ast \bar{\mathbf{K}}.
\end{align}
Making SSMs input-sensitive, Mamba~\cite{mamba,mamba2} is proposed by modifying the following terms:
\begin{align}
    \bar{\mathbf{B}}_t &= \texttt{Linear}_{\textbf{B}}(x_t), \\
    \bar{\mathbf{C}}_t &= \texttt{Linear}_{\textbf{C}}(x_t), \\
    \boldsymbol{\Delta}_t &= \texttt{Softplus}\left(\texttt{Linear}_{\boldsymbol{\Delta}}(x_t)\right).
\end{align}
\noindent \textbf{SPR-Mamba.} 
Given a sequence of images along a trajectory, SPR-Mamba is able to predict the camera pose of an arbitrary query image $I_q$ \textit{\wrt} $I_1$ for an arbitrary sequence length ${I_1, I_2, \ldots\, I_q}$.
Fig.~\ref{fig:spr-mamba} presents the model architecture of the dual-branch SPR-Mamba and sets $q{=}4$ for the illustration purpose.
SPR-Mamba utilizes a frozen DINO~\cite{caron2021dino} to extract image features.
The features are then fed into a local branch to attain frame-by-frame features and a global branch for the query-to-origin features.
Within the local branch, SPR-Mamba first calculates the difference between two consecutive adjacent frames, \ie, $4$ image features result in $3$ feature differences, followed by multiple linear layers.
The auxiliary translation head then outputs $3$ relative translations of the $3$ frame-by-frame relative poses from $4$ images.
The same operation applies to the auxiliary rotation head during training.
Note that the two auxiliary heads are not necessary after training.

In addition to the local branch focusing on the frame-by-frame camera poses, the $4$ DINO-extracted features also go through the global branch for query-to-origin feature learning.
The global branch is stacked by multiple Mamba blocks and finally selects the last hidden state from all outcomes outputted by the last Mamba block since the last one aggregates all information from $I_1$ to $I_q$.
The selected hidden state is then fused with the features outputted by the local branch.
Like the local branch, a translation and a rotation head take the fused feature as input and output the relative query-to-origin camera translation and rotation, respectively.
Since Mamba is capable of doing inference sequentially, SPR-Mamba can continuously handle the upcoming panoramas during inference with linear complexity.

\begin{table*}[ht]
    \caption{Comparison of different models using different paradigms in both \colorbox{Gray!10}{seen} and \colorbox{ForestGreen!20}{unseen} environments on the \textbf{360SPR} dataset. The average median and average mean of Translation Error (TE in meters) and Rotation Error (RE in degrees) are reported. 
    }
    \vskip -1ex
    \label{tab:sota_360spr}
    \renewcommand{\arraystretch}{1.}
    \resizebox{\textwidth}{!}{
    \setlength{\tabcolsep}{4.0pt}
    \begin{tabular}{lcllc|>{\columncolor{Gray!10}}c>{\columncolor{ForestGreen!20}}c>{\columncolor{Gray!10}}c>{\columncolor{ForestGreen!20}}c|>{\columncolor{Gray!10}}c>{\columncolor{ForestGreen!20}}c>{\columncolor{Gray!10}}c>{\columncolor{ForestGreen!20}}c}
    \toprule[1pt]
    \multirow{2}{*}{\textbf{Paradigm}}&\multirow{2}{*}{\textbf{Model}}&\multirow{2}{*}{\textbf{Source}}&\multirow{2}{*}{\textbf{Code}}&\multirow{2}{*}{\textbf{\#Image}}&\multicolumn{4}{c|}{\textbf{Average Median}}&\multicolumn{4}{c}{\textbf{Average Mean}} \\
    {}&{}&{}&{}&{}&{TE (seen)}&{TE (unseen)}&{RE (seen)}&{RE (unseen)}&{TE (seen)}&{TE (unseen)}&{RE (seen)}&{RE (unseen)} \\
    \midrule
    \multirow{6}{*}{APR}&{PoseNet~\cite{kendall2015posenet}}&ICCV&\href{https://github.com/alexgkendall/caffe-posenet}{link}&\faImage[regular]${\times}$1&{10.12\lightgray{$\pm$0.3}}&{30.25\lightgray{$\pm$1.2}}&{10.22\lightgray{$\pm$0.3}}&{47.15\lightgray{$\pm$1.3}}&{10.13\lightgray{$\pm$0.2}}&{29.54\lightgray{$\pm$1.3}}&{10.23\lightgray{$\pm$0.2}}&{46.02\lightgray{$\pm$1.0}} \\
    {}&{NeFeS~\cite{chen2024nefes}}&CVPR&\href{https://github.com/ActiveVisionLab/NeFeS}{link}&\faImage[regular]${\times}$1&{3.29\lightgray{$\pm$0.3}}&{27.88\lightgray{$\pm$0.9}}&{3.40\lightgray{$\pm$0.3}}&{49.01\lightgray{$\pm$0.8}}&{3.20\lightgray{$\pm$0.3}}&{29.04\lightgray{$\pm$1.1}}&{3.22\lightgray{$\pm$0.2}}&{47.51\lightgray{$\pm$0.9}} \\
    {}&{Marepo~\cite{chen2024marepo}}&CVPR&\href{https://github.com/nianticlabs/marepo}{link}&\faImage[regular]${\times}$1&{\textbf{3.22\lightgray{$\pm$0.2}}}&{27.98\lightgray{$\pm$1.1}}&{\textbf{3.31\lightgray{$\pm$0.3}}}&{48.12\lightgray{$\pm$1.1}}&{\textbf{3.13\lightgray{$\pm$0.3}}}&{28.96\lightgray{$\pm$1.2}}&{\textbf{3.02\lightgray{$\pm$0.2}}}&{47.44\lightgray{$\pm$1.2}} \\
    % \cline{2-7}
    \cmidrule(l){2-13}
    {}&{VidLoc~\cite{clark2017vidloc}}&CVPR&\href{https://github.com/futurely/deep-camera-relocalization}{link}&{\faImages[regular]${\times}$5}&{9.23\lightgray{$\pm$0.4}}&{27.44\lightgray{$\pm$1.2}}&{9.62\lightgray{$\pm$0.5}}&{46.99\lightgray{$\pm$1.2}}&{9.43\lightgray{$\pm$0.2}}&{27.45\lightgray{$\pm$1.1}}&{9.17\lightgray{$\pm$0.4}}&{47.33\lightgray{$\pm$1.0}} \\
    {}&{MapNet~\cite{brahmbhatt2018mapnet}}&CVPR&\href{https://github.com/NVlabs/geomapnet}{link}&{\faImages[regular]${\times}$5}&{9.23\lightgray{$\pm$0.2}}&{27.12\lightgray{$\pm$1.2}}&{9.45\lightgray{$\pm$0.6}}&{47.22\lightgray{$\pm$1.3}}&{9.45\lightgray{$\pm$0.2}}&{26.71\lightgray{$\pm$1.2}}&{9.34\lightgray{$\pm$0.3}}&{46.15\lightgray{$\pm$0.7}} \\
    {}&{GL-Net~\cite{xue2020gl-net}}&CVPR&reimpl.&{\faImages[regular]${\times}$5}&{8.61\lightgray{$\pm$0.2}}&{27.45\lightgray{$\pm$1.0}}&{9.31\lightgray{$\pm$0.4}}&{47.01\lightgray{$\pm$0.4}}&{8.91\lightgray{$\pm$0.4}}&{28.44\lightgray{$\pm$1.1}}&{8.89\lightgray{$\pm$0.4}}&{48.21\lightgray{$\pm$1.0}} \\
    \midrule
    \multirow{3}{*}{RPR}&{NN-Net~\cite{laskar2017nn-net}}&ICCVW&\href{https://github.com/AaltoVision/camera-relocalisation?tab=readme-ov-file}{link}&\faImage[regular]${\times}$1&{10.93\lightgray{$\pm$0.3}}&{12.84\lightgray{$\pm$0.4}}&{10.32\lightgray{$\pm$0.2}}&{22.65\lightgray{$\pm$0.2}}&{10.34\lightgray{$\pm$0.3}}&{12.89\lightgray{$\pm$0.2}}&{10.25\lightgray{$\pm$0.5}}&{22.22\lightgray{$\pm$0.4}} \\
    {}&FAR~\cite{rockwell2024far}&CVPR&\href{https://github.com/crockwell/far}{link}&\faImage[regular]${\times}$1&{10.06\lightgray{$\pm$0.3}}&{11.85\lightgray{$\pm$0.3}}&{9.51\lightgray{$\pm$0.3}}&{21.04\lightgray{$\pm$0.2}}&{10.02\lightgray{$\pm$0.3}}&{11.22\lightgray{$\pm$0.4}}&{10.26\lightgray{$\pm$0.4}}&{21.16\lightgray{$\pm$0.5}} \\
    {}&{PanoPose~\cite{tu2024panopose}}&CVPR&reimpl.&\faImage[regular]${\times}$1&{10.01\lightgray{$\pm$0.4}}&{10.91\lightgray{$\pm$0.3}}&{9.02\lightgray{$\pm$0.4}}&{20.01\lightgray{$\pm$0.3}}&{10.23\lightgray{$\pm$0.2}}&{11.03\lightgray{$\pm$0.3}}&{10.12\lightgray{$\pm$0.4}}&{20.55\lightgray{$\pm$0.4}} \\
    \midrule
    \multirow{3}{*}{VO}&{DPVO~\cite{teed2024deep}}&NeurIPS&\href{https://github.com/princeton-vl/DPVO}{link}&\faImage[regular]${\times}$5&{3.88\lightgray{$\pm$0.3}}&{4.02\lightgray{$\pm$0.4}}&{4.12\lightgray{$\pm$0.2}}&{4.38\lightgray{$\pm$0.4}}&{3.71\lightgray{$\pm$0.2}}&{3.92\lightgray{$\pm$0.3}}&{4.35\lightgray{$\pm$0.3}}&{4.44\lightgray{$\pm$0.4}} \\
    {}&{LEAP-VO~\cite{chen2024leapvo}}&CVPR&\href{https://github.com/chiaki530/leapvo}{link}&\faImage[regular]${\times}$5&{3.77\lightgray{$\pm$0.4}}&{3.89\lightgray{$\pm$0.3}}&{4.22\lightgray{$\pm$0.2}}&{4.30\lightgray{$\pm$0.4}}&{3.72\lightgray{$\pm$0.3}}&{3.85\lightgray{$\pm$0.3}}&{4.30\lightgray{$\pm$0.4}}&{4.33\lightgray{$\pm$0.2}} \\
    {}&{XVO~\cite{lai2023xvo}}&ICCV&\href{https://github.com/h2xlab/XVO}{link}&\faImage[regular]${\times}$5&{4.11\lightgray{$\pm$0.3}}&{4.25\lightgray{$\pm$0.3}}&{4.02\lightgray{$\pm$0.3}}&{4.21\lightgray{$\pm$0.3}}&{3.68\lightgray{$\pm$0.4}}&{3.88\lightgray{$\pm$0.4}}&{4.22\lightgray{$\pm$0.2}}&{4.27\lightgray{$\pm$0.4}} \\
    \midrule
    {SPR}&{SPR-Mamba (ours)}&CVPR&\href{https://junweizheng93.github.io/publications/SPR/SPR.html}{link}&{\faImages[regular]${\times}$5}&{3.32\lightgray{$\pm$0.3}}&{\textbf{3.85}\lightgray{$\pm$0.3}}&{3.43\lightgray{$\pm$0.3}}&{\textbf{3.97}\lightgray{$\pm$0.4}}&{3.22\lightgray{$\pm$0.2}}&{\textbf{3.78}\lightgray{$\pm$0.4}}&{3.31\lightgray{$\pm$0.3}}&{\textbf{3.91}\lightgray{$\pm$0.3}} \\
    \bottomrule
    \end{tabular}
    }
    \end{table*}
    
\section{Experiments}
\label{sec:experiments}

\subsection{Datasets}
\label{subsec:datasets}
\noindent\textbf{360SPR.} 
Our new dataset provides over $200{K}$ photorealistic panoramas and the corresponding camera poses in $20{K}$ trajectories distributed in $270$ scenarios.
All panoramas have $1024{\times}2048$ resolution captured at $3$ sensor heights (sweeping~{\includegraphics[width=3mm]{figures/icon_vacuum.png}, quadruped~\includegraphics[width=3mm]{figures/icon_dog.png}, and humanoid~\includegraphics[width=3mm]{figures/icon_robot.png}} robots). \\
\noindent\textbf{360Loc.} 
The dataset~\cite{huang2024360loc} consists of $9.3{K}$ photographic panoramas with corresponding camera poses in $18$ independent trajectories distributed in $4$ scenarios, namely Concourse, Hall, Atrium, and Piatrium. 
All panoramas have a $3072{\times}6144$ resolution captured at a fixed sensor height.

\subsection{Implementation details}
\label{subsec:implementation}
We train the SPR-Mamba model from scratch without any pretraining except for a frozen DINO~\cite{caron2021dino} as the feature extractor.
The SPR-Mamba is trained with an A100 GPU for $150$ epochs.
The AdamW~\cite{loshchilov2017adamw} optimizer is applied with an initial learning rate of $1e^{-4}$. 
The training is warmed by a linear scheduler for the first $10$ epochs followed by a cosine annealing strategy. 
To facilitate the training and inference, we resize the panoramic images to $320{\times}640$ for the 360SPR and $392{\times}770$ for the 360Loc~\cite{huang2024360loc} dataset.
SPR-Mamba is trained with a sequence length of $5$ images and uses the last one as the query image.
Applying a batch size of $8$ results in $40$ images within a batch.
We use L1 loss to supervise every camera pose:
\begin{align}
    L = \alpha \| \hat{\mathbf{t}} - \mathbf{t} \|_1 + \beta \| \hat{\mathbf{q}} - \mathbf{q} \|_1 ,
\end{align}
where $\hat{\mathbf{t}}$ and $\hat{\mathbf{q}}$ are the SPR-Mamba translation and rotation output, $\mathbf{t}$ and $\mathbf{q}$ are the translation and rotation ground-truth, respectively.
$\alpha$ and $\beta$ are two scaling factors balancing the translation and rotation losses.
Empirically, we choose $\alpha=1$ and $\beta=10$.
Following \cite{turkoglu2021relpose}, we parameterize camera rotation as the logarithm of a unit quaternion, equivalent to the axis-angle representation up to scale~\cite{hartley2003multiple}.
This operation avoids the need for additional constraints to ensure a valid rotation.
Logarithmic mapping of the unit quaternion $\mathbf{q} = [u, \mathbf{v}]$ is done via:
\begin{equation}
    \log(\mathbf{q}) =      
    \begin{cases}
      \frac{\mathbf{v}}{\|\mathbf{v}\|_2}\cos^{-1}(u) & \text{if $\|\mathbf{v}\|_2 \neq 0$,} \\
      \mathbf{0} & \text{otherwise,}
    \end{cases}  
\end{equation}
where $u$ and $\mathbf{v}$ are the real and the imaginary part of a unit quaternion, respectively.
The logarithmic form $\mathbf{w} = \log(\mathbf{q})$ can be converted back to a unit quaternion by the exponential mapping, $\exp(\mathbf{w})=[\cos(\|\mathbf{w}\|_2), \frac{\mathbf{w}}{\|\mathbf{w}\|_2}\sin(\|\mathbf{w}\|_2)]$.

\begin{table*}[ht]
    \caption{Comparison of different models using different paradigms in both \colorbox{Gray!10}{seen} and \colorbox{ForestGreen!20}{unseen} environments on the \textbf{360Loc} dataset. The average median and average mean of Translation Error (TE in meters) and Rotation Error (RE in degrees) are reported. 
    } 
    \vskip -1ex
    \label{tab:sota_360loc}
    \renewcommand{\arraystretch}{1.}
    \resizebox{\textwidth}{!}{
    \setlength{\tabcolsep}{4.0pt}
    \begin{tabular}{lcllc|>{\columncolor{Gray!10}}c>{\columncolor{ForestGreen!20}}c>{\columncolor{Gray!10}}c>{\columncolor{ForestGreen!20}}c|>{\columncolor{Gray!10}}c>{\columncolor{ForestGreen!20}}c>{\columncolor{Gray!10}}c>{\columncolor{ForestGreen!20}}c}
    \toprule[1pt]
    \multirow{2}{*}{\textbf{Paradigm}}&\multirow{2}{*}{\textbf{Model}}&\multirow{2}{*}{\textbf{Source}}&\multirow{2}{*}{\textbf{Code}}&\multirow{2}{*}{\textbf{\#Image}}&\multicolumn{4}{c|}{\textbf{Average Median}}&\multicolumn{4}{c}{\textbf{Average Mean}} \\
    {}&{}&{}&{}&{}&{TE (seen)}&{TE (unseen)}&{RE (seen)}&{RE (unseen)}&{TE (seen)}&{TE (unseen)}&{RE (seen)}&{RE (unseen)} \\
    \midrule
    \multirow{6}{*}{APR}&{PoseNet~\cite{kendall2015posenet}}&ICCV&\href{https://github.com/alexgkendall/caffe-posenet}{link}&\faImage[regular]${\times}$1&{8.23\lightgray{$\pm$0.2}}&{28.55\lightgray{$\pm$1.4}}&{8.34\lightgray{$\pm$0.3}}&{45.12\lightgray{$\pm$1.1}}&{8.26\lightgray{$\pm$0.2}}&{27.32\lightgray{$\pm$1.6}}&{8.55\lightgray{$\pm$0.3}}&{46.02\lightgray{$\pm$1.3}} \\
    {}&{NeFeS~\cite{chen2024nefes}}&CVPR&\href{https://github.com/ActiveVisionLab/NeFeS}{link}&\faImage[regular]${\times}$1&{\textbf{1.27\lightgray{$\pm$0.2}}}&{25.78\lightgray{$\pm$0.7}}&{1.12\lightgray{$\pm$0.3}}&{47.12\lightgray{$\pm$0.7}}&{1.21\lightgray{$\pm$0.3}}&{27.12\lightgray{$\pm$1.0}}&{1.12\lightgray{$\pm$0.2}}&{45.63\lightgray{$\pm$0.9}} \\
    {}&{Marepo~\cite{chen2024marepo}}&CVPR&\href{https://github.com/nianticlabs/marepo}{link}&\faImage[regular]${\times}$1&{1.31\lightgray{$\pm$0.2}}&{25.75\lightgray{$\pm$1.2}}&{\textbf{1.11\lightgray{$\pm$0.3}}}&{46.21\lightgray{$\pm$1.1}}&{\textbf{1.17\lightgray{$\pm$0.2}}}&{26.81\lightgray{$\pm$1.4}}&{\textbf{1.03\lightgray{$\pm$0.2}}}&{45.11\lightgray{$\pm$1.5}} \\
    % \cline{2-7}
    \cmidrule(l){2-13}
    {}&{VidLoc~\cite{clark2017vidloc}}&CVPR&\href{https://github.com/futurely/deep-camera-relocalization}{link}&{\faImages[regular]${\times}$5}&{7.36\lightgray{$\pm$0.4}}&{25.33\lightgray{$\pm$1.1}}&{7.72\lightgray{$\pm$0.5}}&{44.39\lightgray{$\pm$1.1}}&{7.66\lightgray{$\pm$0.3}}&{25.13\lightgray{$\pm$1.0}}&{7.17\lightgray{$\pm$0.4}}&{45.19\lightgray{$\pm$1.3}} \\
    {}&{MapNet~\cite{brahmbhatt2018mapnet}}&CVPR&\href{https://github.com/NVlabs/geomapnet}{link}&{\faImages[regular]${\times}$5}&{7.03\lightgray{$\pm$0.4}}&{25.21\lightgray{$\pm$1.3}}&{7.32\lightgray{$\pm$0.7}}&{45.43\lightgray{$\pm$1.3}}&{7.31\lightgray{$\pm$0.4}}&{24.66\lightgray{$\pm$1.1}}&{7.23\lightgray{$\pm$0.3}}&{44.51\lightgray{$\pm$1.1}} \\
    {}&{GL-Net~\cite{xue2020gl-net}}&CVPR&reimpl.&{\faImages[regular]${\times}$5}&{6.45\lightgray{$\pm$0.3}}&{25.72\lightgray{$\pm$1.3}}&{7.44\lightgray{$\pm$0.4}}&{45.23\lightgray{$\pm$1.1}}&{6.94\lightgray{$\pm$0.4}}&{26.65\lightgray{$\pm$1.2}}&{7.01\lightgray{$\pm$0.3}}&{46.11\lightgray{$\pm$1.0}} \\
    \midrule
    \multirow{3}{*}{RPR}&{NN-Net~\cite{laskar2017nn-net}}&ICCVW&\href{https://github.com/AaltoVision/camera-relocalisation?tab=readme-ov-file}{link}&\faImage[regular]${\times}$1&{8.98\lightgray{$\pm$0.2}}&{10.97\lightgray{$\pm$0.3}}&{8.13\lightgray{$\pm$0.4}}&{20.51\lightgray{$\pm$0.2}}&{8.53\lightgray{$\pm$0.4}}&{10.77\lightgray{$\pm$0.4}}&{8.37\lightgray{$\pm$0.3}}&{20.01\lightgray{$\pm$0.4}} \\
    {}&{FAR~\cite{rockwell2024far}}&CVPR&\href{https://github.com/crockwell/far}{link}&\faImage[regular]${\times}$1&{7.99\lightgray{$\pm$0.3}}&{9.35\lightgray{$\pm$0.3}}&{7.42\lightgray{$\pm$0.4}}&{19.01\lightgray{$\pm$0.2}}&{8.02\lightgray{$\pm$0.3}}&{9.13\lightgray{$\pm$0.3}}&{8.16\lightgray{$\pm$0.4}}&{19.05\lightgray{$\pm$0.4}} \\
    {}&{PanoPose~\cite{tu2024panopose}}&CVPR&reimpl.&\faImage[regular]${\times}$1&{7.81\lightgray{$\pm$0.4}}&{9.12\lightgray{$\pm$0.4}}&{7.35\lightgray{$\pm$0.4}}&{18.94\lightgray{$\pm$0.4}}&{7.99\lightgray{$\pm$0.3}}&{9.01\lightgray{$\pm$0.3}}&{8.06\lightgray{$\pm$0.3}}&{19.00\lightgray{$\pm$0.5}} \\
    \midrule
    \multirow{3}{*}{VO}&{DPVO~\cite{teed2024deep}}&NeurIPS&\href{https://github.com/princeton-vl/DPVO}{link}&\faImage[regular]${\times}$5&{2.22\lightgray{$\pm$0.3}}&{2.40\lightgray{$\pm$0.3}}&{1.72\lightgray{$\pm$0.3}}&{1.84\lightgray{$\pm$0.2}}&{2.30\lightgray{$\pm$0.3}}&{2.43\lightgray{$\pm$0.3}}&{1.82\lightgray{$\pm$0.3}}&{2.01\lightgray{$\pm$0.3}} \\
    {}&{LEAP-VO~\cite{chen2024leapvo}}&CVPR&\href{https://github.com/chiaki530/leapvo}{link}&\faImage[regular]${\times}$5&{2.54\lightgray{$\pm$0.2}}&{2.71\lightgray{$\pm$0.3}}&{1.68\lightgray{$\pm$0.2}}&{2.03\lightgray{$\pm$0.3}}&{2.46\lightgray{$\pm$0.3}}&{2.66\lightgray{$\pm$0.3}}&{1.83\lightgray{$\pm$0.3}}&{2.01\lightgray{$\pm$0.3}} \\
    {}&{XVO~\cite{lai2023xvo}}&ICCV&\href{https://github.com/h2xlab/XVO}{link}&\faImage[regular]${\times}$5&{2.43\lightgray{$\pm$0.3}}&{2.56\lightgray{$\pm$0.2}}&{1.80\lightgray{$\pm$0.4}}&{1.99\lightgray{$\pm$0.3}}&{2.51\lightgray{$\pm$0.4}}&{2.69\lightgray{$\pm$0.3}}&{1.68\lightgray{$\pm$0.2}}&{1.92\lightgray{$\pm$0.3}} \\
    \midrule
    {SPR}&{SPR-Mamba (ours)}&CVPR&\href{https://junweizheng93.github.io/publications/SPR/SPR.html}{link}&{\faImages[regular]${\times}$5}&{1.43\lightgray{$\pm$0.3}}&{\textbf{1.94}\lightgray{$\pm$0.3}}&{1.21\lightgray{$\pm$0.2}}&{\textbf{1.44}\lightgray{$\pm$0.2}}&{1.23\lightgray{$\pm$0.3}}&{\textbf{1.87}\lightgray{$\pm$0.3}}&{1.17\lightgray{$\pm$0.3}}&{\textbf{1.28}\lightgray{$\pm$0.2}} \\
    \bottomrule
    \end{tabular}
    }
    \end{table*}
    
\subsection{Quantitative Results}
To benchmark the proposed SPR task, we conduct experiments on two panoramic datasets, \ie, our 360SPR and the 360Loc~\cite{huang2024360loc} dataset. All models are trained three times for a fair comparison. The average median and mean of translation error (TE in meters) and rotation error (RE in degrees) among all scenes with uncertainty are reported. 

\noindent \textbf{Results on 360SPR.} 
Table~\ref{tab:sota_360spr} shows the comparison results of different paradigms. 
In the \colorbox{ForestGreen!20}{unseen} setting, we leave out $15$ of $270$ scenes as unknown during training and use them for testing. Image-based APR~\cite{kendall2015posenet,chen2024nefes,chen2024marepo} and sequence-based APR~\cite{clark2017vidloc,brahmbhatt2018mapnet,xue2020gl-net} models have much lower generalizability in unseen environments. 
For example, Marepo~\cite{chen2024marepo} obtains the average median of TE/RE in $27{m}$/$48^{\circ}$ even though it achieves the best performance in seen environments.
The reason for the poor generalization is that models in the APR paradigm learn scene-specific features to predict the absolute camera pose.
Therefore, they fail to predict poses in novel scenes.
On the other hand, RPR models generalize better than APR models since they focus on the relative features between frames which are less related to the scenes.
VO models can generalize well in unseen environments.
However, compared to SPR, VO performs worse due to the unavoidable accumulated drift.
Our SPR-Mamba in the SPR paradigm significantly outperforms all APR, RPR and VO models in unknown environments, yielding the best performance (the average median of TE/RE in $\bf{3.85{m}/3.97^{\circ}}$). The error reduction of TE/RE are over $\bf{23{m}/43^{\circ}{\downarrow}}$ compared to APR, and $\bf{7{m}/16^{\circ}{\downarrow}}$ compared to RPR. The SPR paradigm disentangles the coordinate system from specific scenes and predicts the relative pose between the query and origin frame by learning the scene-agnostic features along trajectories. These results show that our SPR paradigm and model generalize better in unknown environments. 
In the \colorbox{Gray!10}{seen} setting, $20\%$ trajectories of training scenes are excluded for testing. Surprisingly, the performance of our SPR-Mamba is on par with the best image-based APR method Marepo~\cite{chen2024marepo}.

\noindent \textbf{Results on 360Loc.} 
Table~\ref{tab:sota_360loc} shows the comparison on the 360Loc~\cite{huang2024360loc} dataset. In the \colorbox{ForestGreen!20}{unseen} environments, we perform cross-validation, in which $1$ of $4$ scenes is excluded as unknown and for testing, while the remaining $3$ scenes for training. The same operation repeats until all $4$ scenes are tested. We then average the results of $4$ scenes. Consistent with the results on 360SPR, all APR models generalize poorly in unseen scenes on the 360Loc~\cite{huang2024360loc} dataset. RPR models showcase the superiority in unseen environments compared to both image-based and sequence-based APR models. Although VO models perform better than APR and RPR models in unknown environments, they underperform models in the SPR paradigm. Our SPR-Mamba outperforms all models, obtaining the best performance (the average median of TE/RE in $\bf{1.94{m}/1.44^{\circ}}$) with over $\bf{7{m}/17^{\circ}{\downarrow}}$ error reduction of the RPR paradigm. In the \colorbox{Gray!10}{seen} testing, the data splitting follows the official 360Loc~\cite{huang2024360loc} dataset. The performance gap between SPR-Mamba and Marepo~\cite{chen2024marepo} is marginal with $0.12{m}$ TE and $0.1^{\circ}$ RE. The remarkable performance on the real-world panoramic dataset proves the effectiveness of our SPR-Mamba in both seen and unseen environments. 

\begin{figure}[t]
    \centering
    \begin{subfigure}{0.22\textwidth}
        \centering
        \includegraphics[width=\linewidth]{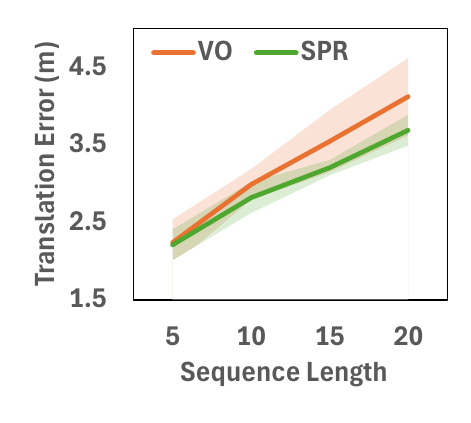}
    \end{subfigure}
    \begin{subfigure}{0.22\textwidth}
        \centering
        \includegraphics[width=\linewidth]{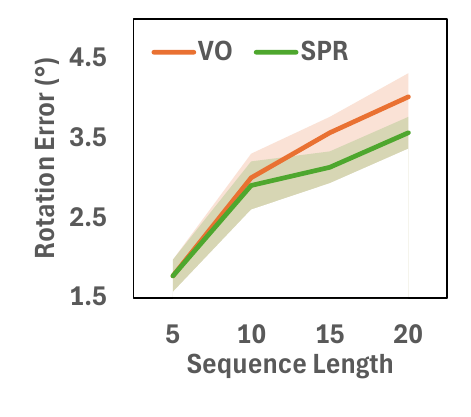}
    \end{subfigure}
    \vskip -2ex
    \caption{Comparison of VO and SPR paradigms for unknown scenes with different sequence lengths of 360Loc~\cite{huang2024360loc} dataset. }
    \vskip -2ex
    \label{fig:vo_comparison}
\end{figure}

\subsection{Ablation Study}
\noindent \textbf{Comparison with VO.} 
Fig.~\ref{fig:vo_comparison} shows the comparison between VO and SPR paradigms.
Compared with VO, SPR predicts the camera pose relative to the first origin frame instead of the previous frame while VO predicts the relative pose between two consecutive adjacent frames.
Since VO predicts the current pose based on the previous one, the accumulated drift in open trajectories is unavoidable. In contrast, our SPR paradigm directly estimates the pose according to the origin, avoiding the need for loop closure detection to eliminate the drift. 
We train TSformer-VO~\cite{franccani2023tsformervo} in both paradigms on the 360Loc~\cite{huang2024360loc} dataset with different sequence lengths, from $5$ to $20$ images.
The model is trained with $3$ scenes and evaluated with the last remaining scene in a cross-validation manner.
We train the model three times and report the average median of TE and RE with standard deviation in shadow in Fig.~\ref{fig:vo_comparison}.
It can be observed that the TE and RE of TSformer-VO~\cite{franccani2023tsformervo} become larger as the sequence becomes longer in both SPR and VO settings. However, the performance gap widens as the sequence length increases since the SPR paradigm does not experience drift, whereas the VO paradigm does. The result confirms that our SPR paradigm can better handle long sequences.

\noindent\textbf{Component Ablation.} 
To further analyze the effectiveness of different components of the proposed SPR-Mamba, we ablate the local branch, global branch, and auxiliary training heads. The average median of TE and RE in unknown environments on the 360SPR dataset are reported in Table~\ref{tab:ablation_spr-mamba}. 
\circled{1} Without the global branch, SPR-Mamba degenerates into a simple model consisting of several linear layers, predicting the consecutive adjacent frames like VO. Apart from the aforementioned variant, \circled{2} SPR-Mamba without the local branch and the auxiliary regression heads cannot achieve satisfying results, either.
The model only focuses on the query-to-origin relative features, neglecting the local relative features which help improve overall model performance.
Compared with the previous variant, this model has a performance gain over $5{m}/5^{\circ}$ for TE and RE, respectively.
\circled{3} SPR-Mamba with a local and global branch but without the auxiliary regression heads, achieves even better performance since the model is able to capture the local relative features even though there is no supervision for the relative poses of the consecutive adjacent frames.
\circled{4} SPR-Mamba with all components, performs the best.
In addition to the local and global feature learning, the auxiliary heads provide more supervision via their loss functions for the local branch, which further benefits the global feature learning through gradient and backpropagation.

\begin{table}[!t]
\begin{center}
\vskip -1ex
\caption{Ablation study of SPR-Mamba components on 360SPR dataset. Translation/Rotation Errors (TE/RE in $m$/{\textdegree}) are reported.}
\vskip -2ex
\label{tab:ablation_spr-mamba}
\setlength{\tabcolsep}{3mm}
\resizebox{\columnwidth}{!}{
\renewcommand{\arraystretch}{1.2}
    \begin{tabular}{lccccc}
    \toprule[1pt]
    {\textbf{Model}} & {\textbf{Aux. Heads}} & {\textbf{Local Branch}} & {\textbf{Global Branch}} & {\textbf{TE}} & {\textbf{RE}} \\
    \midrule
    \circled{1}~{SPR-Mamba} & {\includegraphics[width=4mm]{figures/icon_checkmark.png}} & {\includegraphics[width=4mm]{figures/icon_checkmark.png}} & {\includegraphics[width=4mm]{figures/icon_crossmark.png}} & {10.33\lightgray{$\pm$1.3}} & {10.68\lightgray{$\pm$1.4}}  \\
    \circled{2}~{SPR-Mamba} & {\includegraphics[width=4mm]{figures/icon_crossmark.png}} & {\includegraphics[width=4mm]{figures/icon_crossmark.png}} & {\includegraphics[width=4mm]{figures/icon_checkmark.png}} & {4.72\lightgray{$\pm$0.7}} & {5.11\lightgray{$\pm$0.5}}  \\
    \circled{3}~{SPR-Mamba} & {\includegraphics[width=4mm]{figures/icon_crossmark.png}} & {\includegraphics[width=4mm]{figures/icon_checkmark.png}} & {\includegraphics[width=4mm]{figures/icon_checkmark.png}} & {4.32\lightgray{$\pm$0.6}} & {4.67\lightgray{$\pm$0.5}}  \\
    \circled{4}~{SPR-Mamba} & {\includegraphics[width=4mm]{figures/icon_checkmark.png}} & {\includegraphics[width=4mm]{figures/icon_checkmark.png}} & {\includegraphics[width=4mm]{figures/icon_checkmark.png}} & {\textbf{3.85}\lightgray{$\pm$0.3}} & {\textbf{3.97}\lightgray{$\pm$0.4}}  \\
    \bottomrule[1pt]
    \end{tabular}
}
\end{center}
\vskip -2ex
\end{table}

\noindent\textbf{Cross-data Evaluation.}
We conduct the cross-evaluation experiment between 360Loc~\cite{huang2024360loc} and 360SPR datasets.
Table~\ref{tab:dataset_cross-evaluation} demonstrates the average median of TE and RE for unseen environments of both datasets.
When evaluated on 360Loc~\cite{huang2024360loc}, SPR-Mamba trained on 360SPR achieves comparable results to those trained on 360Loc~\cite{huang2024360loc}.
The performance gap of $0.2{m}/0.23^{\circ}$ on TE/RE is neglectable.
In contrast, SPR-Mamba trained on 360Loc~\cite{huang2024360loc} is not able to perform well on the 360SPR dataset because 360SPR is more diverse and challenging.
It's difficult for a model to generalize well when trained on a dataset with a single sensor height, similar scenes and limited panoramic data.
The performance gap, in this case, reaches $2.17{m}/2.14^{\circ}$ on TE/RE.
This ablation study shows the value of creating a large-scale panoramic dataset like 360SPR with over $200{K}$ panoramas and $270$ scenes at $3$ different sensor heights, making a model more robust to real-world cases.

\begin{table}[tb]
    \centering
    \captionof{table}{Evaluation between datasets and between sensor heights. Translation/Rotation Errors (TE/RE in $m$/{\textdegree}) are reported.}
    \label{tab:cross-evaluation}

    \vskip -2ex
    \includegraphics[width=\linewidth]{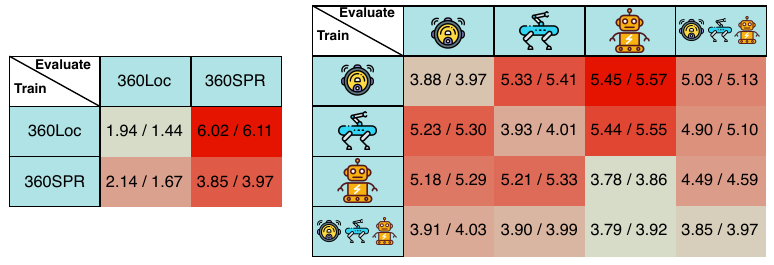}
    
    \begin{subtable}[t]{0.4\linewidth}
        \centering
        \vskip-2ex
        \caption{Evaluation cross datasets.}
        \label{tab:dataset_cross-evaluation}
    \end{subtable}%\hfill
     \begin{subtable}[t]{0.6\linewidth}
        \centering
        \vskip-2ex
        \caption{Evaluation cross sensor heights.} \label{tab:height_cross-evaluation}
    \end{subtable}%\hfill
\vskip -2ex
\end{table}

\noindent\textbf{Cross-sensor Evaluation.}
Apart from the cross-dataset evaluation, we also ablate the influence of different sensor heights in unknown environments.
As shown in Table~\ref{tab:height_cross-evaluation}, we train SPR-Mamba model on $4$ splits of 360SPR, namely the data only at $0.1{m}$~\includegraphics[width=3mm]{figures/icon_vacuum.png}, data only at $0.5{m}$~\includegraphics[width=3mm]{figures/icon_dog.png}, data only at $1.7{m}$~\includegraphics[width=3mm]{figures/icon_robot.png}, and a mixture of all heights~\includegraphics[width=3mm]{figures/icon_vacuum.png}~\includegraphics[width=3mm]{figures/icon_dog.png}~\includegraphics[width=3mm]{figures/icon_robot.png}.
It can be observed that the domain gap exists across different heights, \eg, SPR-Mamba trained on pure $1.7{m}$ data achieves a $3.78{m}$ average median of translation error evaluated at the same height, while the error increases by $1.4{m}{\uparrow}$ when evaluated at the pure $0.1{m}$ data.
Compared with the training at a single height, SPR-Mamba showcases excellent robustness when trained on the complete 360SPR dataset, achieving the best/second-best results in all splits.
This ablation study proves the necessity of creating a dataset with different sensor heights for accurate and robust camera pose regression.

\section{Conclution}
\label{sec:conclusion}
We propose Scene-agnostic Pose Regression (SPR) to enhance APR generalization and reduce RPR’s database dependency.
To explore SPR, we create a large-scale panoramic dataset (360SPR) with over $200{K}$ photorealistic panoramic images and $3.6{M}$ pinhole images in $270$ scenes. To address SPR, a dual-branch SPR-Mamba model is constructed with an SSM-based mechanism, which showcases its superiority in the camera pose regression task in unseen environments with a $7{m}/16^{\circ}$ error degradation in translation and rotation compared with APR and RPR models.
We hope this work has the potential to advance the field of camera pose regression, providing generalized visual localization in unknown scenes. \\
\noindent \textbf{Limitations and Future Work.} Although the SPR paradigm is capable of predicting camera poses in unknown environments, the poses are relative to the origin frame.
No information about the absolute poses is available. Besides, the ability of SPR-Mamba to handle image distortions in panoramas will be further improved in our future work.

\clearpage
\section*{Acknowledgments}
This work was supported in part by the Ministry of Science, Research and the Arts of Baden-W\"urttemberg (MWK) through the Cooperative Graduate School Accessibility through AI-based Assistive Technology (KATE) under Grant BW6-03, in part by Karlsruhe House of Young Scientists (KHYS), in part by the Helmholtz Association Initiative and Networking Fund on the HAICORE@KIT and HOREKA@KIT partition, in part by the National Natural Science Foundation of China (No.~62473139), and in part by the National Key RD Program under Grant~2022YFB4701400. 

%%%%%%%%% REFERENCES
% \clearpage
{\small
\bibliographystyle{ieeenat_fullname}
\bibliography{main}
}

\clearpage

\appendix

\section{Dataset Construction}

We create a large-scale panoramic dataset, \textbf{360SPR}, for not only the Scene-agnostic Pose Regression but also other visual localization tasks, such as Absolute Pose Regression and Relative Pose Regression. Leveraging the Habitat simulator~\cite{savva2019habitat1,szot2021habitat2,puig2023habitat3} powered by HM3D~\cite{ramakrishnan2021hm3d} and Matterport3D~\cite{chang2017matterport3d} datasets, we sample over $3.6{M}$ pinhole images with corresponding camera poses and depth images distributed in $270$ different scenes.
$180$ scenes come from HM3D~\cite{ramakrishnan2021hm3d} and  the rest scenes are from Matterport3D~\cite{chang2017matterport3d}.
For the sake of obtaining panoramas, we use the same stitching tool as Matterport3D~\cite{chang2017matterport3d} to stitch pinholes into panoramas.

\begin{figure}[tb]
    \centering
    \includegraphics[width=\linewidth]{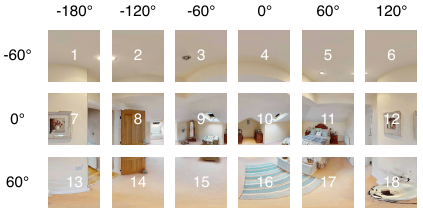}
    %\vskip -2ex
    \caption{One panorama is stitched by $18$ pinholes with $6$ headings and $3$ elevations. The numbers in white represent the image indices in the sequence.}
    \label{sup_fig:image_collection}
    %\vskip -2ex
\end{figure}

As shown in Fig.~\ref{sup_fig:image_collection}, for every sample point in the trajectories, we collect images with $6$ headings and $3$ elevations, resulting in $18$ pinhole images.
Each pinhole image has a $60{^\circ}$ horizontal and vertical field of view in $512{\times}512$.
As for the heading and elevation, they are also $60{^\circ}$, resulting in $360{^\circ}$ horizontal and $180{^\circ}$ vertical field-of-view stitched panoramic images.
Referring to the camera pose of the $i$-th panorama along a trajectory, we leverage the face direction from the $(i{-}1)$-th sample point pointing to the $i$-th sample point, which is also the pose of the $10$-th pinhole image in the pinhole image sequence of the $i$-th panorama.
We also randomly add a heading offset ranging from ${-}60{^\circ}$ to $60{^\circ}$ to the panoramic camera poses for diversity.
To enable high-quality panoramic images, three inspectors manually checked all samples in the form of cross-validation. 
The whole cleaning process took more than $300$ hours.

As for the trajectory selection, we randomly select two points as the starting and destination points within a navigable area of a scene.
Then we calculate the shortest path between the two points using the Dijkstra~\cite{dijkstra2022dijkstra} algorithm.
Since the 360Loc dataset~\cite{huang2024360loc} doesn't consider different sampling intervals and sensor heights, it's difficult to satisfy the need for robust and accurate spatial awareness in various real-world applications.
To this end, we sample trajectories in different lengths with varying sampling intervals between sampling points along the path.
The trajectory length in 360SPR varies from $3{m}$ to $20{m}$ and the number of panoramas in one trajectory varies from $5$ to $20$.

Moreover, three different robot heights with a sampling ratio of $1{:}1{:}2$ are also taken into account, \ie, sweeping~({\includegraphics[width=3mm]{figures/icon_vacuum.png}), quadruped~(\includegraphics[width=3mm]{figures/icon_dog.png}), and humanoid~(\includegraphics[width=3mm]{figures/icon_robot.png}}) robots.
Note that one trajectory corresponds to one robot's height rather than a mixture of three different heights.

\begin{table}[!t]
\begin{center}
\caption{Model specifications of SPR-Mamba.}
\vskip -2ex
\label{sup_tab:model_details}
\setlength{\tabcolsep}{1mm}
\resizebox{\columnwidth}{!}{
\renewcommand{\arraystretch}{1.2}
    \begin{tabular}{lcccccc}
    \toprule[1pt]
    {\textbf{Branch}} & {\textbf{Block}} & {\textbf{Block Num.}} & {\textbf{Input Dim.}} & {\textbf{Hidden Dim.}} & {\textbf{Hidden States}} & {\textbf{Output Dim.}} \\
    \midrule
    Feature Extractor & DINOv2s & 1 & - & - & - & 384 \\
    Local Branch & Linear & 12 & 384 & 768 & - & 384 \\
    Global Branch & Mamba & 12 & 384 & 768 & 16 & 384 \\
    Translation Head & Linear & 1 & 384 & - & - & 3 \\
    Rotation Head & Linear & 1 & 384 & - & - & 3 \\
    \bottomrule[1pt]
    \end{tabular}
}
\end{center}
\vskip -2ex
\end{table}
\section{More Implementation Details}
We train the SPR-Mamba model from scratch without any pretraining except for a frozen DINO~\cite{caron2021dino} as the feature extractor.
The SPR-Mamba is trained with an A100 GPU for $150$ epochs.
The AdamW~\cite{loshchilov2017adamw} optimizer is applied with an initial learning rate of $1e^{-4}$. The training is warmed by a linear scheduler for the first $10$ epochs followed by a cosine annealing strategy. To facilitate the training and inference, we resize the panoramic images to $320{\times}640$ for the 360SPR and $392{\times}770$ for the 360Loc dataset~\cite{huang2024360loc}.
SPR-Mamba is trained with a sequence length of $5$ images and uses the last one as the query image.
Applying a batch size of $8$ results in $40$ images within a batch.

Table~\ref{sup_tab:model_details} lists the model specification of SPR-Mamba.
We utilize DINOv2s~\cite{caron2021dino,oquab2023dinov2} as the feature extractor.
As for the Linear layer in the local branch, we stack $12$ Linear layers where the hidden layer dimension is twice as large as the input and output dimensions.
We also stack $12$ Mamba blocks in the global branch where the expand ratio is $2$ with $16$ hidden states.
The Mamba~\cite{mamba,mamba2} blocks are tailored to handle more global contextual information, with the expansion ratio helping to enlarge the model capacity and improve overall performance.

\section{More Quantitative Results}
\begin{table*}[ht]
    \caption{Comparison of different models using different paradigms in both \colorbox{Gray!10}{seen} and \colorbox{ForestGreen!20}{unseen} environments on the \textbf{360SPR} dataset. The average median and average mean of Translation Error (TE in meters) and Rotation Error (RE in degrees) are reported. 
    }
    \label{sup_tab:sota_360spr}
    \renewcommand{\arraystretch}{1.}
    \resizebox{\textwidth}{!}{
    \setlength{\tabcolsep}{4.0pt}
    \begin{tabular}{lcllc|>{\columncolor{Gray!10}}c>{\columncolor{ForestGreen!20}}c>{\columncolor{Gray!10}}c>{\columncolor{ForestGreen!20}}c|>{\columncolor{Gray!10}}c>{\columncolor{ForestGreen!20}}c>{\columncolor{Gray!10}}c>{\columncolor{ForestGreen!20}}c}
    \toprule[1pt]
    \multirow{2}{*}{\textbf{Paradigm}}&\multirow{2}{*}{\textbf{Model}}&\multirow{2}{*}{\textbf{Source}}&\multirow{2}{*}{\textbf{Code}}&\multirow{2}{*}{\textbf{\#Image}}&\multicolumn{4}{c|}{\textbf{Average Median}}&\multicolumn{4}{c}{\textbf{Average Mean}} \\
    {}&{}&{}&{}&{}&{TE (seen)}&{TE (unseen)}&{RE (seen)}&{RE (unseen)}&{TE (seen)}&{TE (unseen)}&{RE (seen)}&{RE (unseen)} \\
    \midrule
    \multirow{3}{*}{APR}&{AnchorPoint~\cite{saha2018anchorpoint}}&BMVC&\href{https://github.com/Soham0/Improved-Visual-Relocalization}{link}&\faImage[regular]${\times}$1&{10.11\lightgray{$\pm$0.4}}&{29.44\lightgray{$\pm$0.9}}&{10.11\lightgray{$\pm$0.5}}&{46.66\lightgray{$\pm$1.2}}&{10.14\lightgray{$\pm$0.2}}&{28.23\lightgray{$\pm$0.8}}&{10.51\lightgray{$\pm$0.2}}&{47.13\lightgray{$\pm$1.3}} \\
    {}&{MS-Transformer~\cite{shavit2021ms-transformer}}&ICCV&\href{https://github.com/yolish/c2f-ms-transformer}{link}&\faImage[regular]${\times}$1&{10.22\lightgray{$\pm$0.4}}&{30.35\lightgray{$\pm$1.2}}&{10.11\lightgray{$\pm$0.3}}&{47.65\lightgray{$\pm$0.9}}&{10.16\lightgray{$\pm$0.3}}&{29.37\lightgray{$\pm$1.1}}&{10.65\lightgray{$\pm$0.2}}&{48.32\lightgray{$\pm$1.3}} \\
    {}&{DFNet~\cite{chen2022dfnet}}&ECCV&\href{https://github.com/ActiveVisionLab/DFNet}{link}&\faImage[regular]${\times}$1&{3.87\lightgray{$\pm$0.4}}&{28.35\lightgray{$\pm$0.7}}&{3.69\lightgray{$\pm$0.6}}&{47.84\lightgray{$\pm$1.0}}&{3.92\lightgray{$\pm$0.2}}&{28.33\lightgray{$\pm$0.7}}&{3.75\lightgray{$\pm$0.2}}&{47.53\lightgray{$\pm$1.2}} \\
    \midrule
    \multirow{3}{*}{RPR}&{RelocNet~\cite{balntas2018relocnet}}&ECCV&\href{https://relocnet.active.vision/dataset.html}{link}&\faImage[regular]${\times}$1&{10.55\lightgray{$\pm$0.5}}&{12.45\lightgray{$\pm$0.2}}&{10.21\lightgray{$\pm$0.4}}&{21.42\lightgray{$\pm$0.2}}&{10.33\lightgray{$\pm$0.3}}&{11.42\lightgray{$\pm$0.3}}&{10.64\lightgray{$\pm$0.5}}&{21.19\lightgray{$\pm$0.4}} \\
    {}&{Ess-Net~\cite{zhou2020essnet}}&ICRA&\href{https://github.com/GrumpyZhou/visloc-relapose}{link}&\faImage[regular]${\times}$1&{10.12\lightgray{$\pm$0.3}}&{12.54\lightgray{$\pm$0.5}}&{9.87\lightgray{$\pm$0.4}}&{21.44\lightgray{$\pm$0.4}}&{10.76\lightgray{$\pm$0.3}}&{11.52\lightgray{$\pm$0.3}}&{10.21\lightgray{$\pm$0.2}}&{21.48\lightgray{$\pm$0.3}} \\
    {}&{Relpose-GNN~\cite{turkoglu2021relpose}}&3DV&\href{https://github.com/nianticlabs/relpose-gnn}{link}&\faImage[regular]${\times}$1&{10.19\lightgray{$\pm$0.4}}&{11.92\lightgray{$\pm$0.4}}&{9.62\lightgray{$\pm$0.4}}&{21.27\lightgray{$\pm$0.2}}&{10.26\lightgray{$\pm$0.2}}&{11.44\lightgray{$\pm$0.6}}&{10.51\lightgray{$\pm$0.5}}&{21.33\lightgray{$\pm$0.6}} \\
    \midrule
    {SPR}&{SPR-Mamba (ours)}&CVPR&\href{https://junweizheng93.github.io/publications/SPR/SPR.html}{link}&{\faImages[regular]${\times}5$}&{\textbf{3.32}\lightgray{$\pm$0.3}}&{\textbf{3.85}\lightgray{$\pm$0.3}}&{\textbf{3.43}\lightgray{$\pm$0.3}}&{\textbf{3.97}\lightgray{$\pm$0.4}}&{\textbf{3.22}\lightgray{$\pm$0.2}}&{\textbf{3.78}\lightgray{$\pm$0.4}}&{\textbf{3.31}\lightgray{$\pm$0.3}}&{\textbf{3.91}\lightgray{$\pm$0.3}} \\
    \bottomrule
    \end{tabular}
    }
    \end{table*}
    
\subsection{More Results on 360SPR}
In addition to the comparison with other state-of-the-art baselines in the main paper, we provide more quantitative comparisons in this section.
Table~\ref{sup_tab:sota_360spr} compares SPR-Mamba with more baselines in both seen and unseen environments on the 360SPR dataset.
It can be observed that SPR-Mamba surpasses other methods, achieving an average reduction of $8\text{m}/17{^\circ}\downarrow$ in median translation and rotation errors in unseen environments.
This result is also consistent with the result in our main paper.

\begin{table*}[ht]
    \caption{Comparison of different models using different paradigms in both \colorbox{Gray!10}{seen} and \colorbox{ForestGreen!20}{unseen} environments on the \textbf{360Loc} dataset. The average median and average mean of Translation Error (TE in meters) and Rotation Error (RE in degrees) are reported. 
    }
    \label{sup_tab:sota_360loc}
    \renewcommand{\arraystretch}{1.}
    \resizebox{\textwidth}{!}{
    \setlength{\tabcolsep}{4.0pt}
    \begin{tabular}{lcllc|>{\columncolor{Gray!10}}c>{\columncolor{ForestGreen!20}}c>{\columncolor{Gray!10}}c>{\columncolor{ForestGreen!20}}c|>{\columncolor{Gray!10}}c>{\columncolor{ForestGreen!20}}c>{\columncolor{Gray!10}}c>{\columncolor{ForestGreen!20}}c}
    \toprule[1pt]
    \multirow{2}{*}{\textbf{Paradigm}}&\multirow{2}{*}{\textbf{Model}}&\multirow{2}{*}{\textbf{Source}}&\multirow{2}{*}{\textbf{Code}}&\multirow{2}{*}{\textbf{\#Image}}&\multicolumn{4}{c|}{\textbf{Average Median}}&\multicolumn{4}{c}{\textbf{Average Mean}} \\
    {}&{}&{}&{}&{}&{TE (seen)}&{TE (unseen)}&{RE (seen)}&{RE (unseen)}&{TE (seen)}&{TE (unseen)}&{RE (seen)}&{RE (unseen)} \\
    \midrule
    \multirow{3}{*}{APR}&{AnchorPoint~\cite{saha2018anchorpoint}}&BMVC&\href{https://github.com/Soham0/Improved-Visual-Relocalization}{link}&\faImage[regular]${\times}$1&{8.16\lightgray{$\pm$0.3}}&{27.25\lightgray{$\pm$1.3}}&{8.15\lightgray{$\pm$0.3}}&{44.52\lightgray{$\pm$1.4}}&{8.27\lightgray{$\pm$0.2}}&{26.12\lightgray{$\pm$1.1}}&{8.35\lightgray{$\pm$0.2}}&{45.11\lightgray{$\pm$1.7}} \\
    {}&{MS-Transformer~\cite{shavit2021ms-transformer}}&ICCV&\href{https://github.com/yolish/c2f-ms-transformer}{link}&\faImage[regular]${\times}$1&{8.31\lightgray{$\pm$0.2}}&{28.45\lightgray{$\pm$1.5}}&{8.27\lightgray{$\pm$0.2}}&{45.76\lightgray{$\pm$1.2}}&{8.33\lightgray{$\pm$0.1}}&{27.31\lightgray{$\pm$1.2}}&{8.44\lightgray{$\pm$0.3}}&{46.41\lightgray{$\pm$1.6}} \\
    {}&{DFNet~\cite{chen2022dfnet}}&ECCV&\href{https://github.com/ActiveVisionLab/DFNet}{link}&\faImage[regular]${\times}$1&{1.85\lightgray{$\pm$0.4}}&{26.22\lightgray{$\pm$0.8}}&{1.77\lightgray{$\pm$0.6}}&{45.62\lightgray{$\pm$1.1}}&{1.95\lightgray{$\pm$0.2}}&{26.44\lightgray{$\pm$0.8}}&{1.95\lightgray{$\pm$0.3}}&{45.89\lightgray{$\pm$1.0}} \\
    \midrule
    \multirow{3}{*}{RPR}&{RelocNet~\cite{balntas2018relocnet}}&ECCV&\href{https://relocnet.active.vision/dataset.html}{link}&\faImage[regular]${\times}$1&{8.65\lightgray{$\pm$0.3}}&{10.73\lightgray{$\pm$0.3}}&{8.01\lightgray{$\pm$0.3}}&{19.51\lightgray{$\pm$0.2}}&{8.62\lightgray{$\pm$0.3}}&{9.98\lightgray{$\pm$0.4}}&{8.24\lightgray{$\pm$0.4}}&{19.55\lightgray{$\pm$0.3}} \\
    {}&{Ess-Net~\cite{zhou2020essnet}}&ICRA&\href{https://github.com/GrumpyZhou/visloc-relapose}{link}&\faImage[regular]${\times}$1&{8.57\lightgray{$\pm$0.2}}&{10.43\lightgray{$\pm$0.4}}&{7.92\lightgray{$\pm$0.2}}&{19.67\lightgray{$\pm$0.2}}&{8.51\lightgray{$\pm$0.2}}&{9.74\lightgray{$\pm$0.2}}&{8.15\lightgray{$\pm$0.4}}&{19.32\lightgray{$\pm$0.4}} \\
    {}&{Relpose-GNN~\cite{turkoglu2021relpose}}&3DV&\href{https://github.com/nianticlabs/relpose-gnn}{link}&\faImage[regular]${\times}$1&{8.02\lightgray{$\pm$0.3}}&{9.98\lightgray{$\pm$0.2}}&{7.77\lightgray{$\pm$0.4}}&{19.45\lightgray{$\pm$0.4}}&{8.22\lightgray{$\pm$0.4}}&{9.82\lightgray{$\pm$0.4}}&{8.01\lightgray{$\pm$0.2}}&{19.02\lightgray{$\pm$0.2}} \\
    \midrule
    {SPR}&{SPR-Mamba (ours)}&CVPR&\href{https://junweizheng93.github.io/publications/SPR/SPR.html}{link}&{\faImages[regular]${\times}5$}&{\textbf{1.43}\lightgray{$\pm$0.3}}&{\textbf{1.94}\lightgray{$\pm$0.3}}&{\textbf{1.21}\lightgray{$\pm$0.2}}&{\textbf{1.44}\lightgray{$\pm$0.2}}&{\textbf{1.23}\lightgray{$\pm$0.3}}&{\textbf{1.87}\lightgray{$\pm$0.3}}&{\textbf{1.17}\lightgray{$\pm$0.3}}&{\textbf{1.28}\lightgray{$\pm$0.2}} \\
    \bottomrule
    \end{tabular}
    }
    \end{table*}
    
\begin{table}[th]
    \caption{
    Results in unseen environments on pinhole datasets \textbf{7Scenes} and \textbf{360SPR pinhole subset}. 
    }
    \label{sup_tab:pinholes}
    \renewcommand{\arraystretch}{1.}
    \resizebox{\columnwidth}{!}{
    \setlength{\tabcolsep}{4.0pt}
    \begin{tabular}{lcc|cc|cc}
    \toprule[1pt]
    \multirow{2}{*}{\textbf{Paradigm}}&\multirow{2}{*}{\textbf{Model}}&\multirow{2}{*}{\textbf{Source}}&\multicolumn{2}{c|}{\textbf{7Scenes (Pinhole)}}&\multicolumn{2}{c}{\textbf{360SPR (Pinhole)}} \\
    {}&{}&{}&{TE~(m)$\downarrow$}&{RE~(\textdegree)$\downarrow$}&{TE~(m)$\downarrow$}&{RE~(\textdegree)$\downarrow$} \\
    \midrule
    {APR}&{Marepo~\cite{chen2024marepo}
    }&CVPR&{2.02\lightgray{$\pm$0.3}}&{3.54\lightgray{$\pm$0.3}}&{9.53\lightgray{$\pm$0.3}}&{11.31\lightgray{$\pm$0.3}} \\
    \midrule
    {RPR}&{FAR~\cite{rockwell2024far}}&CVPR&{1.83\lightgray{$\pm$0.3}}&{3.22\lightgray{$\pm$0.4}}&{9.03\lightgray{$\pm$0.2}}&{10.98\lightgray{$\pm$0.2}} \\
    \midrule
    \multirow{3}{*}{VO}&{DPVO~\cite{teed2024deep}}&NeurIPS&{0.66\lightgray{$\pm$0.3}}&{1.54\lightgray{$\pm$0.3}}&{4.33\lightgray{$\pm$0.4}}&{5.21\lightgray{$\pm$0.3}} \\
    {}&{LEAP-VO~\cite{chen2024leapvo}}&CVPR&{0.73\lightgray{$\pm$0.3}}&{1.77\lightgray{$\pm$0.3}}&{4.47\lightgray{$\pm$0.4}}&{5.51\lightgray{$\pm$0.4}} \\
    {}&{XVO~\cite{lai2023xvo}}&ICCV&{0.70\lightgray{$\pm$0.1}}&{1.69\lightgray{$\pm$0.4}}&{4.55\lightgray{$\pm$0.3}}&{5.33\lightgray{$\pm$0.4}} \\
    \midrule
    \multirow{2}{*}{SPR}&{SPR-Transformer (ours)}&CVPR&{0.44\lightgray{$\pm$0.3}}&{1.23\lightgray{$\pm$0.4}}&{4.04\lightgray{$\pm$0.4}}&{5.01\lightgray{$\pm$0.2}} \\
    {}&{SPR-Mamba (ours)}&CVPR&{\textbf{0.40}\lightgray{$\pm$0.3}}&{\textbf{1.21}\lightgray{$\pm$0.3}}&{\textbf{3.96}\lightgray{$\pm$0.3}}&{\textbf{4.89}\lightgray{$\pm$0.2}} \\
    \bottomrule
    \end{tabular}
    }
\end{table}
    
\subsection{More Results on 360Loc}
We also compare SPR-Mamba with more baselines in both seen and unseen environments on the 360Loc dataset~\cite{huang2024360loc}.
The results are reported in Table~\ref{sup_tab:sota_360loc}.
It can be observed that SPR-Mamba surpasses other methods, achieving an average reduction of $8\text{m}/18{^\circ}\downarrow$ in median translation and rotation errors in unseen environments. Models trained in the APR paradigm are still not able to work in unknown environments.
The results on the 360SPR and 360Loc~\cite{huang2024360loc} datasets prove the effectiveness of our proposed SPR paradigm in predicting accurate and robust camera poses in unknown environments.

\subsection{Results on Pinhole Datasets}
Table~\ref{sup_tab:pinholes} showcases the results on two pinhole datasets, namely 7Scenes~\cite{glocker20137scenes,shotton20137scenes} and 200K-pinhole subset of our 360SPR.
Thanks to our model design and SPR paradigm, SPR-Mamba performs consistently well on pinhole datasets, as compared to APR, RPR, and VO.
Moreover, we also compare our model with different architectures in Table~\ref{sup_tab:pinholes}, namely Transformer-based and Mamba-based models.
Besides the lower computational complexity, Mamba achieves better performance.

\begin{table*}[ht]
    % \tiny
    % \setlength{\abovecaptionskip}{0pt}
    % \setlength{\belowcaptionskip}{0pt}
    \caption{Ablation study of SPR-Mamba at different sensor heights (0.1m \includegraphics[width=3mm]{figures/icon_vacuum.png}, 0.5m \includegraphics[width=3mm]{figures/icon_dog.png}, 1.7m \includegraphics[width=3mm]{figures/icon_robot.png}) in both \colorbox{Gray!10}{seen} and \colorbox{ForestGreen!20}{unseen} environments on the \textbf{360SPR} dataset. The average median and mean of Translation Error (TE in meters) and Rotation Error (RE in degrees) are reported. %TE and RE stand for Translation Error and Rotation Error, respectively.
    }
    \label{sup_tab:sensor_height}
    % \begin{center}
    \renewcommand{\arraystretch}{1.}
    \resizebox{\textwidth}{!}{
    \setlength{\tabcolsep}{8.0pt}
    \begin{tabular}{c|>{\columncolor{Gray!10}}c>{\columncolor{ForestGreen!20}}c>{\columncolor{Gray!10}}c>{\columncolor{ForestGreen!20}}c|>{\columncolor{Gray!10}}c>{\columncolor{ForestGreen!20}}c>{\columncolor{Gray!10}}c>{\columncolor{ForestGreen!20}}c}
    \toprule[1pt]
    \multirow{2}{*}{\textbf{Height}}&\multicolumn{4}{c|}{\textbf{Average Median}}&\multicolumn{4}{c}{\textbf{Average Mean}} \\
    {}&{TE (seen)}&{TE (unseen)}&{RE (seen)}&{RE (unseen)}&{TE (seen)}&{TE (unseen)}&{RE (seen)}&{RE (unseen)} \\
    \midrule
    {\includegraphics[width=3mm]{figures/icon_vacuum.png}}&{3.33\lightgray{$\pm$0.3}}&{3.88\lightgray{$\pm$0.2}}&{\textbf{3.24}\lightgray{$\pm$0.2}}&{3.97\lightgray{$\pm$0.3}}&{3.56\lightgray{$\pm$0.3}}&{3.77\lightgray{$\pm$0.4}}&{3.48\lightgray{$\pm$0.3}}&{\textbf{3.68}\lightgray{$\pm$0.3}} \\
    {\includegraphics[width=3mm]{figures/icon_dog.png}}&{3.29\lightgray{$\pm$0.2}}&{{3.93}\lightgray{$\pm$0.2}}&{3.31\lightgray{$\pm$0.2}}&{{4.01}\lightgray{$\pm$0.2}}&{3.65\lightgray{$\pm$0.3}}&{{3.88}\lightgray{$\pm$0.4}}&{3.44\lightgray{$\pm$0.3}}&{{3.74}\lightgray{$\pm$0.3}} \\
    {\includegraphics[width=3mm]{figures/icon_robot.png}}&{3.32\lightgray{$\pm$0.3}}&{{3.78}\lightgray{$\pm$0.3}}&{3.27\lightgray{$\pm$0.3}}&{{3.86}\lightgray{$\pm$0.3}}&{3.35\lightgray{$\pm$0.3}}&{{3.76}\lightgray{$\pm$0.2}}&{3.33\lightgray{$\pm$0.3}}&{{3.85}\lightgray{$\pm$0.2}} \\
    {\includegraphics[width=3mm]{figures/icon_vacuum.png}\includegraphics[width=3mm]{figures/icon_dog.png}}&{3.11\lightgray{$\pm$0.2}}&{\textbf{3.46}\lightgray{$\pm$0.3}}&{3.33\lightgray{$\pm$0.4}}&{{3.78}\lightgray{$\pm$0.4}}&{3.44\lightgray{$\pm$0.2}}&{\textbf{3.69}\lightgray{$\pm$0.3}}&{\textbf{3.21}\lightgray{$\pm$0.2}}&{{3.88}\lightgray{$\pm$0.3}} \\
    {\includegraphics[width=3mm]{figures/icon_vacuum.png}\includegraphics[width=3mm]{figures/icon_robot.png}}&{\textbf{3.10}\lightgray{$\pm$0.2}}&{{3.55}\lightgray{$\pm$0.3}}&{3.67\lightgray{$\pm$0.3}}&{\textbf{3.69}\lightgray{$\pm$0.3}}&{3.40\lightgray{$\pm$0.2}}&{{3.99}\lightgray{$\pm$0.4}}&{3.24\lightgray{$\pm$0.3}}&{{3.87}\lightgray{$\pm$0.4}} \\
    {\includegraphics[width=3mm]{figures/icon_dog.png}\includegraphics[width=3mm]{figures/icon_robot.png}}&{3.62\lightgray{$\pm$0.3}}&{{3.66}\lightgray{$\pm$0.3}}&{3.51\lightgray{$\pm$0.3}}&{{3.88}\lightgray{$\pm$0.3}}&{3.30\lightgray{$\pm$0.3}}&{{3.82}\lightgray{$\pm$0.2}}&{3.42\lightgray{$\pm$0.3}}&{{3.77}\lightgray{$\pm$0.4}} \\
    {\includegraphics[width=3mm]{figures/icon_vacuum.png}\includegraphics[width=3mm]{figures/icon_dog.png}\includegraphics[width=3mm]{figures/icon_robot.png}}&{3.32\lightgray{$\pm$0.3}}&{{3.85}\lightgray{$\pm$0.3}}&{3.43\lightgray{$\pm$0.3}}&{{3.97}\lightgray{$\pm$0.4}}&{\textbf{3.22}\lightgray{$\pm$0.2}}&{{3.78}\lightgray{$\pm$0.4}}&{3.31\lightgray{$\pm$0.3}}&{{3.91}\lightgray{$\pm$0.3}} \\
    \bottomrule
    \end{tabular}
    }
    % \end{center}
    \end{table*}
    
\begin{table*}[ht]
    \caption{Ablation study of SPR-Mamba and TSformer-VO with different sequence lengths in both \colorbox{Gray!10}{seen} and \colorbox{ForestGreen!20}{unseen} environments on the \textbf{360Loc} dataset. The average median and mean of Translation Error (TE in meters) and Rotation Error (RE in degrees) are reported. 
    }
    \label{sup_tab:seq_len}
    \renewcommand{\arraystretch}{1.}
    \resizebox{\textwidth}{!}{
    \setlength{\tabcolsep}{6.0pt}
    \begin{tabular}{lc|>{\columncolor{Gray!10}}c>{\columncolor{ForestGreen!20}}c>{\columncolor{Gray!10}}c>{\columncolor{ForestGreen!20}}c|>{\columncolor{Gray!10}}c>{\columncolor{ForestGreen!20}}c>{\columncolor{Gray!10}}c>{\columncolor{ForestGreen!20}}c}
    \toprule[1pt]
    \multirow{2}{*}{\textbf{Model}}&\multirow{2}{*}{\textbf{\#Image}}&\multicolumn{4}{c|}{\textbf{Average Median}}&\multicolumn{4}{c}{\textbf{Average Mean}} \\
    {}&{}&{TE (seen)}&{TE (unseen)}&{RE (seen)}&{RE (unseen)}&{TE (seen)}&{TE (unseen)}&{RE (seen)}&{RE (unseen)} \\
    \midrule
    {TSformer-VO~\cite{franccani2023tsformervo}}&\faImage[regular]${\times}$5&{2.07\lightgray{$\pm$0.3}}&{2.21\lightgray{$\pm$0.3}}&{1.59\lightgray{$\pm$0.3}}&{1.78\lightgray{$\pm$0.3}}&{2.11\lightgray{$\pm$0.3}}&{2.32\lightgray{$\pm$0.2}}&{1.55\lightgray{$\pm$0.2}}&{1.81\lightgray{$\pm$0.3}} \\
    {SPR-Mamba}&\faImage[regular]${\times}$5&{\textbf{1.43}\lightgray{$\pm$0.3}}&{\textbf{1.94}\lightgray{$\pm$0.3}}&{\textbf{1.21}\lightgray{$\pm$0.2}}&{\textbf{1.44}\lightgray{$\pm$0.2}}&{\textbf{1.23}\lightgray{$\pm$0.3}}&{\textbf{1.87}\lightgray{$\pm$0.3}}&{\textbf{1.17}\lightgray{$\pm$0.3}}&{\textbf{1.28}\lightgray{$\pm$0.2}} \\
    \midrule
    {TSformer-VO~\cite{franccani2023tsformervo}}&\faImage[regular]${\times}$10&{2.56\lightgray{$\pm$0.2}}&{2.82\lightgray{$\pm$0.3}}&{2.77\lightgray{$\pm$0.2}}&{2.91\lightgray{$\pm$0.2}}&{2.61\lightgray{$\pm$0.3}}&{2.79\lightgray{$\pm$0.3}}&{2.81\lightgray{$\pm$0.2}}&{2.92\lightgray{$\pm$0.2}} \\
    {SPR-Mamba}&\faImage[regular]${\times}$10&{\textbf{2.07}\lightgray{$\pm$0.2}}&{\textbf{2.20}\lightgray{$\pm$0.2}}&{\textbf{2.21}\lightgray{$\pm$0.2}}&{\textbf{2.43}\lightgray{$\pm$0.3}}&{\textbf{2.15}\lightgray{$\pm$0.3}}&{\textbf{2.22}\lightgray{$\pm$0.3}}&{\textbf{2.25}\lightgray{$\pm$0.4}}&{\textbf{2.51}\lightgray{$\pm$0.2}} \\
    \midrule
    {TSformer-VO~\cite{franccani2023tsformervo}}&\faImage[regular]${\times}$15&{3.05\lightgray{$\pm$0.2}}&{3.21\lightgray{$\pm$0.2}}&{3.01\lightgray{$\pm$0.3}}&{3.14\lightgray{$\pm$0.3}}&{3.14\lightgray{$\pm$0.3}}&{3.33\lightgray{$\pm$0.4}}&{3.12\lightgray{$\pm$0.2}}&{3.20\lightgray{$\pm$0.2}} \\
    {SPR-Mamba}&\faImage[regular]${\times}$15&{\textbf{2.44}\lightgray{$\pm$0.3}}&{\textbf{2.62}\lightgray{$\pm$0.3}}&{\textbf{2.42}\lightgray{$\pm$0.2}}&{\textbf{2.65}\lightgray{$\pm$0.2}}&{\textbf{2.50}\lightgray{$\pm$0.3}}&{\textbf{2.68}\lightgray{$\pm$0.3}}&{\textbf{2.52}\lightgray{$\pm$0.3}}&{\textbf{2.70}\lightgray{$\pm$0.3}} \\
    \midrule
    {TSformer-VO~\cite{franccani2023tsformervo}}&\faImage[regular]${\times}$20&{3.44\lightgray{$\pm$0.2}}&{3.69\lightgray{$\pm$0.3}}&{3.33\lightgray{$\pm$0.3}}&{3.57\lightgray{$\pm$0.3}}&{3.58\lightgray{$\pm$0.3}}&{3.75\lightgray{$\pm$0.2}}&{3.50\lightgray{$\pm$0.2}}&{3.69\lightgray{$\pm$0.2}} \\
    {SPR-Mamba}&\faImage[regular]${\times}$20&{\textbf{2.65}\lightgray{$\pm$0.3}}&{\textbf{2.89}\lightgray{$\pm$0.2}}&{\textbf{2.71}\lightgray{$\pm$0.3}}&{\textbf{2.93}\lightgray{$\pm$0.2}}&{\textbf{2.60}\lightgray{$\pm$0.3}}&{\textbf{2.90}\lightgray{$\pm$0.2}}&{\textbf{2.88}\lightgray{$\pm$0.3}}&{\textbf{3.10}\lightgray{$\pm$0.2}} \\
    \bottomrule[1pt]
    \end{tabular}
    }
    \end{table*}
    
\begin{table}[t]
    % \tiny
    % \setlength{\abovecaptionskip}{0pt}
    % \setlength{\belowcaptionskip}{0pt}
    \caption{
    %Comparison of different models using different paradigms in unseen environments on the \textbf{7Scenes} dataset and the \textbf{200K-pinhole 360SPR} subset. 
    Results in unseen environments on pinhole \textbf{360SPR pinhole subset} and panoramic \textbf{360SPR} with less overlap. %The average median of Translation Error (TE in meters) and Rotation Error (RE in degrees) are reported. 
    %TE and RE stand for Translation Error and Rotation Error, respectively.
    }
    \label{sup_tab:overlap}
    % \begin{center}
    \renewcommand{\arraystretch}{1.}
    \resizebox{\columnwidth}{!}{
    \setlength{\tabcolsep}{4.0pt}
    \begin{tabular}{lcc|cc|cc}
    \toprule[1pt]
    \multirow{2}{*}{\textbf{Paradigm}}&\multirow{2}{*}{\textbf{Model}}&\multirow{2}{*}{\textbf{Source}}&\multicolumn{2}{c|}{\textbf{360SPR (Pinhole)}}&\multicolumn{2}{c}{\textbf{360SPR (Panoramic)}} \\
    {}&{}&{}&{TE~(m)$\downarrow$}&{RE~(\textdegree)$\downarrow$}&{TE~(m)$\downarrow$}&{RE~(\textdegree)$\downarrow$} \\
    \midrule
    \multirow{3}{*}{VO}&{DPVO~\cite{teed2024deep}}&NeurIPS&{5.53\lightgray{$\pm$0.3}}&{6.33\lightgray{$\pm$0.2}}&{5.04\lightgray{$\pm$0.3}}&{5.55\lightgray{$\pm$0.4}} \\
    {}&{LEAP-VO~\cite{chen2024leapvo}}&CVPR&{5.47\lightgray{$\pm$0.2}}&{6.27\lightgray{$\pm$0.3}}&{5.02\lightgray{$\pm$0.3}}&{5.78\lightgray{$\pm$0.3}} \\
    {}&{XVO~\cite{lai2023xvo}}&ICCV&{5.55\lightgray{$\pm$0.2}}&{6.35\lightgray{$\pm$0.3}}&{5.09\lightgray{$\pm$0.3}}&{5.65\lightgray{$\pm$0.3}} \\
    \midrule
    {SPR}&{SPR-Mamba (ours)}&CVPR&{\textbf{4.33}\lightgray{$\pm$0.3}}&{\textbf{5.47}\lightgray{$\pm$0.2}}&{\textbf{4.03}\lightgray{$\pm$0.2}}&{\textbf{4.23}\lightgray{$\pm$0.3}} \\
    \bottomrule
    \end{tabular}
    }
    % \end{center}
\end{table}
    
\subsection{Results of Less Overlap}
To test less overlapping cases, we further conduct experiments by removing a few frames within a sequence.
Table~\ref{sup_tab:overlap} lists results in unseen environments on 200K-pinhole subset of 360SPR and panoramic 360SPR with less overlap.
Our method consistently outperforms other VO methods on pinhole and panoramic datasets.
It's worth noting that the performance on the panoramic dataset is better than the one on the pinhole dataset since panoramas provide more overlap and visual information compared to the pinhole images.

\subsection{Ablation Study}
\noindent\textbf{Ablation on sensor height.}
We perform a comprehensive ablation study to evaluate the impact of varying sensor heights on the performance of our SPR-Mamba model.
Table~\ref{sup_tab:sensor_height} presents a detailed comparison of the model's performance under the combination of three distinct sensor height configurations: $0.1$ meters \includegraphics[width=3mm]{figures/icon_vacuum.png}, $0.5$ meters \includegraphics[width=3mm]{figures/icon_dog.png}, and $1.7$ meters \includegraphics[width=3mm]{figures/icon_robot.png}.
Unlike the ablation study of cross-sensor evaluation in the main paper, the model is trained and evaluated at the same height with a sequence length of $5$ images in this ablation study.
The results demonstrate that SPR-Mamba maintains consistently high performance across all sensor height combinations.
This consistency underscores the robustness of our model.
Such findings highlight SPR-Mamba's potential for deployment in diverse environments and scenarios.

\noindent\textbf{Ablation on sequence length.}
We conduct an ablation study on image sequence length.
Different from the ablation on VO comparison in the main paper, where we use the same model in two different paradigms, namely VO and SPR, we leverage two models in the same SPR paradigm in this ablation study.
The analysis presented in the main paper investigates the differences between VO and SPR across various sequence lengths.
In contrast, this ablation study focuses specifically on exploring the performance differences among models in the SPR paradigm when subjected to different sequence lengths ranging from $5$ to $20$.
Table~\ref{sup_tab:seq_len} showcases the ablation results.
Note that since TSformer-VO~\cite{franccani2023tsformervo} and SPR-Mamba are both trained and evaluated in the SPR paradigm, there is no accumulated drift in this ablation study.
It can be observed that the translation and rotation errors increase as the image sequence becomes longer.
This phenomenon happens both in TSformer-VO~\cite{franccani2023tsformervo} and SPR-Mamba.
However, our SPR-Mamba consistently outperforms TSformer-VO~\cite{franccani2023tsformervo} in all sequence-length settings in both seen and unseen environments.
This remarkable superiority proves that although extended sequence lengths have the potential to degrade model performance in the SPR paradigm, this challenge is not insurmountable. By employing thoughtful architectural design, as demonstrated by SPR-Mamba, it is possible to effectively alleviate the negative impact of long sequences.

\section{Samples from 360SPR}
\begin{figure*}[t!]
    \centering
    \includegraphics[width=0.99\linewidth]{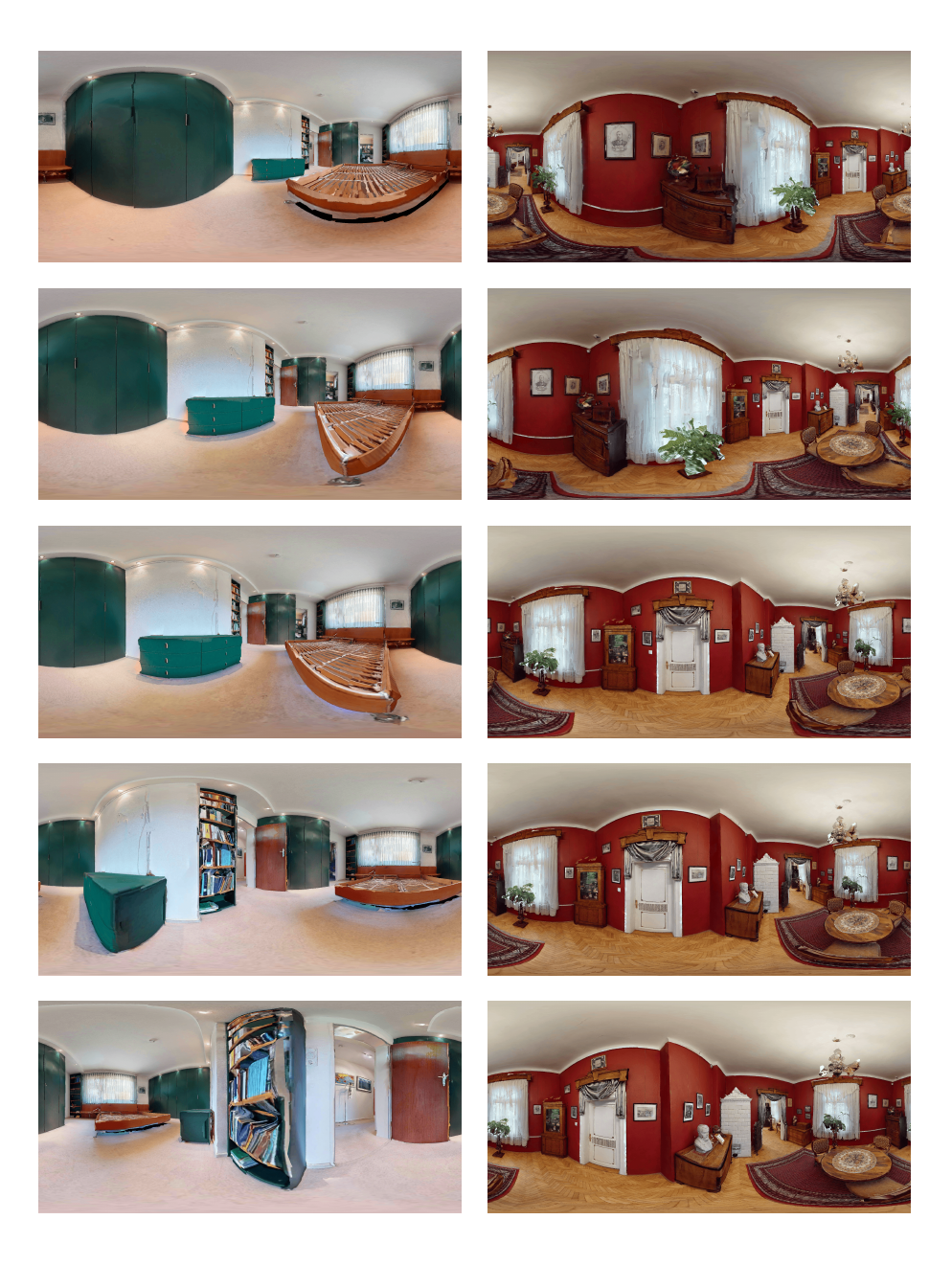}
    \vskip -6ex
    \caption{Samples from 360SPR. We respectively pick $5$ images from $2$ trajectories in $2$ scenes for illustration.}
    \label{sup_fig:360spr_samples}
\end{figure*}
When using pinhole images, substantial changes in the viewpoint, \eg, $180{^\circ}$ rotation, may result in insufficient overlap, which is important for Relative Pose Regression and Scene-agnostic Pose Regression.
In contrast, panoramas guarantee sufficient overlap and similarity since they provide $360{^\circ}$ field of view.
Fig.~\ref{sup_fig:360spr_samples} showcases some data samples from the 360SPR dataset.
We respectively pick $5$ images from $2$ trajectories in $2$ scenes for illustration.
It can be observed that two consecutive adjacent panoramas provide sufficient overlap and similarity to train an accurate and robust pose regression model.

\section{Limitation and Future Work}
While Scene-agnostic Pose Regression is capable of predicting precise camera poses in unfamiliar environments, these poses are defined relative to the origin frame, with no information provided regarding the absolute poses.
360SPR is a large-scale panoramic dataset for visual localization tasks.
It contains panoramas, pinholes, and depth images with camera poses captured at $3$ different sensor heights distributed in $270$ scenes.
In order to satisfy the need for other computer vision tasks beyond visual localization, it's necessary to enrich the 360SPR dataset with more modalities, \eg, segmentation maps.
Although panoramas provide more visual cues compared with pinholes, image distortion occurs due to the spherical projection.
We plan to enhance SPR-Mamba's ability to manage image distortions in panoramas in future work.
Furthermore, given the rapid advancement of Large Language Models(LLMs), exploring the integration of multi-modal LLMs presents an increasingly promising and exciting direction for future research.

\end{document}